\documentclass[conferemce]{IEEEtran}
\usepackage{graphicx}
\usepackage{amsthm}
\usepackage{epsfig}
\usepackage{latexsym}
\usepackage{amsfonts}
\usepackage{here}
\usepackage{rawfonts}
\usepackage[latin1]{inputenc}
\usepackage[T1]{fontenc}
\usepackage{calc}
\usepackage{capitalgreekitalic}
\usepackage{url}
\usepackage{enumerate}
\usepackage{color}
\usepackage[tbtags]{amsmath}
\usepackage{amssymb}
\usepackage{upref}
\usepackage{epic,eepic}
\usepackage{times}
\usepackage{dsfont}
\usepackage{comment}
\usepackage{cite}
\usepackage{bbm} 
\usepackage{multirow}

\usepackage{threeparttable}

\usepackage{dsfont}

\newcounter{step}
\newlength{\totlinewidth}
  {\end{list}%
  \rule{\linewidth}{1pt}}
\newcounter{substep}

  {\end{list}}

\newlength{\aligntop}
\newlength{\alignbot}
 

\IEEEoverridecommandlockouts

\usepackage{algorithm}
\usepackage{algorithmic}

\makeatletter
\newcommand\semihuge{\@setfontsize\semihuge{19.3}{25}}
\makeatother

\makeatletter
\newcommand\semismall{\@setfontsize\semihuge{12.4}{15}}
\makeatother

\usepackage{subfigure}

\usepackage{tablefootnote}

\begin{document}

\title{{\color{black}Distributed Learning in Wireless Networks: Recent Progress and Future Challenges}\vspace*{-0em}}

\author{{Mingzhe Chen}, \emph{Member, IEEE}, Deniz G\"und\"uz, \emph{Senior Member, IEEE}, Kaibin Huang, \emph{Fellow, IEEE},\\ Walid Saad, \emph{Fellow, IEEE}, Mehdi Bennis, \emph{Fellow, IEEE}, Aneta Vulgarakis Feljan, \emph{Member, IEEE},\\ and H. Vincent Poor, \emph{Life Fellow, IEEE}, \vspace*{-2.2em}\\ 
\thanks{M. Chen and H. V. Poor are with the Department of Electrical and Computer Engineering, Princeton University, Princeton, NJ, 08544, USA, Emails: \protect{mingzhec@princeton.edu}, \protect{poor@princeton.edu}.}
\thanks{D. G\"und\"uz is with the Department of Electrical and Electronic Engineering, Imperial College London, London, SW7 2AZ, UK, Email: \protect{d.gunduz@imperial.ac.uk}.}
\thanks{K. Huang is with the Department of Electrical \& Electronic Engineering, The University of Hong Kong, Hong Kong, Email: \protect{huangkb@eee.hku.hk}.}
\thanks{W. Saad is with the Wireless@VT, Bradley Department of Electrical and Computer Engineering, Virginia Tech, Blacksburg, VA, 24060, USA, Email: \protect{walids@vt.edu}.}
\thanks{M. Bennis is with the Department of Communications Engineering, University of Oulu, FI-90014, Oulu Finland, Email: \protect{mehdi.bennis@oulu.fi}.
}
\thanks{A. V. Feljan is with Ericsson Research, Stockholm, Sweden, Email: \protect{aneta.vulgarakis@ericsson.com}.
}
 }
\maketitle
%

\begin{abstract}
The next-generation of wireless networks will enable many machine learning (ML) tools and applications to efficiently analyze various types of data collected by edge devices for inference, autonomy, and decision making purposes. However, due to resource constraints, delay limitations, and privacy challenges, edge devices cannot offload their entire collected datasets to a cloud server for centrally training their ML models or inference purposes. To overcome these challenges, distributed learning and inference techniques have been proposed as a means to enable edge devices to collaboratively train ML models without raw data exchanges, thus reducing the communication overhead and latency as well as improving data privacy. However, deploying distributed learning over wireless networks faces several challenges including the uncertain wireless environment (e.g., dynamic channel and interference), limited wireless resources (e.g., transmit power and radio spectrum), and hardware resources (e.g., computational power). This paper provides a comprehensive study of how distributed learning can be efficiently and effectively deployed over wireless edge networks. We present a detailed overview of several emerging distributed learning paradigms, including federated learning, federated distillation, distributed inference, and multi-agent reinforcement learning. For each learning framework, we first introduce the motivation for deploying it over wireless networks. Then, we present a detailed literature review on the use of communication techniques for its efficient deployment. We then introduce an illustrative example to show how to optimize wireless networks to improve its performance. Finally, we introduce future research opportunities.   In a nutshell, this paper provides a holistic set of guidelines on how to deploy a broad range of distributed learning frameworks over real-world wireless communication networks.  
\end{abstract}

\section{Introduction}
Over the past five years, the field of machine learning (ML) witnessed a major paradigm shift from the so-called ``big data'' paradigm, in which large volumes of data are collected and processed at a central cloud, towards a ``small data'' paradigm~\cite{saad2019vision}, in which a set of distributed agents or devices must process their data in-situ at the edge of a wireless or computing system. This paradigm shift meant that classical centralized ML approaches that require large training datasets to effectively perform inference tasks are no longer applicable. In contrast, there is a major need for novel \emph{distributed learning} solutions that can collaboratively perform inference and learning without the need to exchange local datasets. Such distributed learning solutions must, in essence, be cognizant of the multi-agent, distributed nature of the emerging small data-based applications and systems. The real-world use of this paradigm shift towards distributed learning can be exemplified in the context of Internet of Things (IoT) as well as connected autonomy (e.g., connected vehicles or drones). In such systems, each device collects its own, individualized dataset, that is often private, and, collectively, the devices must be able to train a model while overcoming their local data scarcity. In such scenarios, exchange of raw data is often undesirable (due to privacy reasons) or, in some cases, even infeasible (due to communication and computing constraints).

\subsection{Challenges of Deploying Distributed Learning}
Ensuring a successful evolution towards distributed learning requires overcoming several important challenges. The first key challenge is the fact that distributed learning must occur without exchange of raw data due to privacy reasons. This, in turn, gives rise to an interesting trade-off between maintaining privacy and maximizing accuracy, when deciding on what information to exchange among distributed agents. The second challenge stems from the fact that one common denominator across all such distributed systems is the need to perform training and inference of the ML model over a wireless system, such as a cellular system or a wireless local area network (WLAN). As such, the very nature of the wireless system -- an uncertain environment that is highly stochastic due to key factors such as interference and fading -- will now introduce new impediments to the learning performance. For example, in \cite{chen2019joint}, it has been shown that wireless bit errors and delay can significantly impact the convergence and accuracy of distributed learning over real-world wireless systems. Hence, facilitating the deployment of distributed learning entails a need for jointly accounting for wireless factors in the design. Similarly, it is shown in \cite{Abad:ICASSP20} that the wireless network architecture can have a significant impact on the convergence speed of the learning algorithm. A third, related challenge, is the fact that the radio spectrum bandwidth is highly limited. As such, exchanging some of the information that could be needed for distributed learning, such as model parameters or model updates, can consume significant bandwidth, and, thus, there is a need for communication-efficient techniques to perform distributed learning. Closely tied with these communication challenge is the fourth challenge pertaining to computing. Distributed learning, in all of its flavors, requires efficient ways to perform computations, both over-the-air \cite{Amiri:TSP:20, AirComp_f2} and at the edge. The delay and efficiency associated with distributed computing for ML over wireless networks will directly impact the quality of learning. Finally, distributed learning over wireless systems can involve important network optimization tasks that are performed, collaboratively, by a group of agents. Therefore, there is a need for effective approaches that enable multiple agents to solve complex optimization problems in a distributed way.

Research efforts aimed at solving these challenges led to the emergence of many important distributed learning frameworks in the past few years. Chief among them is the popular \emph{federated learning (FL)} \cite{47976} framework that enables a group of agents to collaboratively execute a common learning task by exchanging only their model parameters, rather than their raw data. By doing so, FL provides a first-level of protection for the users data and, as shown~\cite{47976}, it could yield highly accurate inference. Following the seminal work in \cite{47976}, a broad range of flavors of FL were developed, addressing many of the aforementioned challenges. Meanwhile, to perform distributed optimization, the framework of multi-agent reinforcement learning (MARL)~\cite{4445757} has gained rapid popularity. By combining the concept of reinforcement learning with deep neural networks (NNs) as well as distributed multi-agent control, one can enable a group of agents to solve a set of distributed optimization problems, without the need to rely on global information or without significant exchange of data. Naturally, MARL itself faces many challenges including the need to establish convergence, optimality, and real-time operation. Both FL and MARL will have to operate over large-scale wireless systems and, as such, they are subject to many of the wireless-related challenges that we previously outlined.

 \begin{table*}
 {\color{black}
\centering
  \newcommand{\tabincell}[2]{\begin{tabular}{@{}#1@{}}#2\end{tabular}}
\renewcommand\arraystretch{1}
 \caption{
    \vspace*{-0.1em}Comparison of This Work With Existing Survey and Tutorial Papers. {Here, ``FL", ``DI", ``FD", ``ML'', and ``MARL"  refer to federated learning, distributed inference, federated distillation, machine learning, and multi-agent reinforcement learning, respectively. }  }\label{ta:existingworks}\vspace*{-0.6em}
\centering
\begin{tabular}{|c | c|c| c| c |c |c |c |c |}
\hline
 \multicolumn{1}{|c|}{\multirow{2}{1cm}{\textbf{Existing Works}}} &  \multicolumn{5}{|c|}{\multirow{1}{*}{\textbf{ Key ML Tools }}} &  \multicolumn{3}{|c|}{\multirow{1}{*}{\textbf{ Key Applications}}} \\
  \cline{2-9}
 &\textbf{FL}&\textbf{Beyond FL}&\textbf{DI}&\textbf{FD}&\textbf{MARL} &\textbf{ML for Communications}&\textbf{Communications for ML} &\textbf{Security and Privacy Issues} \\
\hline
\cite{9141214,9205981,9085259,8994206,yang2021federated}&$\surd$&& & & &$\surd$&&\\
\cite{8970161,9084352,9247530,9199786,9357490}&$\surd$ &  & && &&$\surd$&\\
\cite{9060868}&$\surd$ &  & && &$\surd$&$\surd$&$\surd$\\
\cite{9311906,9311931}   && $\surd$ &&&&&$\surd$& \\
\cite{9048613,9220181}& $\surd$ & && &&&&$\surd$\\
\cite{9311939}&$\surd$&&&&&&$\surd$& \\
\cite{8865093}&$\surd$& &&$\surd$&&&   &\\
\cite{8714026}& && & &$\surd$&$\surd$&& \\
\cite{Gunduz:CL:20}& $\surd$&&$\surd$ &$\surd$ &&&$\surd$& \\
Our work&$\surd$ &$\surd$ &$\surd$  &$\surd$ &$\surd$&&$\surd$&\\
\hline
\end{tabular}
}
\end{table*}
\subsection{Previous Works}
In this broad area of distributed learning over wireless networks, a number of surveys and tutorials have recently appeared \cite{9141214,9205981,9085259,8994206,yang2021federated,8970161,9084352,9247530,9199786, 9357490,9060868, 9311906,9311931,9048613,9220181,9311939,8865093,8714026,Gunduz:CL:20}. In \cite{9141214,9205981,9085259,8994206,yang2021federated}, the authors provided comprehensive surveys for the use of FL for communication networks. The authors in \cite{8970161,9084352,9247530,9199786,9357490} introduced the use of communication techniques to improve the performance of FL algorithms that are implemented over wireless networks. The authors in \cite{9311906,9311931} introduced novel distributed learning algorithms based on FL and discussed several open research directions toward practical realization of their solutions. In \cite{9048613,9220181}, the authors provide a survey on the security and privacy challenges of FL. The authors in \cite{9311939} introduced the use of communication techniques to improve the convergence speed and achieve accurate training and inference. In \cite{8865093} the authors introduced FL and federated distillation for communication-efficient learning. The authors in \cite{8714026} explained the application of reinforcement learning for solving wireless communication problems. In \cite{Gunduz:CL:20}, the authors provided an introduction for federated edge learning, distributed inference, and distributed training, and pointed out the challenges of their deployment over wireless networks. However, these prior surveys \cite{9141214,9205981,9085259,8994206,yang2021federated,8970161,9084352,9247530,9199786, 9357490, 9311906,9311931,9048613,9220181,9311939,8865093,8714026,Gunduz:CL:20,9060868} are often restricted in scope (e.g., focused only on FL \cite{9141214,9205981,9085259,8994206,yang2021federated,8970161,9084352,9247530,9199786, 9357490,9048613,9220181,9311939,9060868} or only on MARL \cite{8714026}), do not discuss the major intertwining between wireless systems and distributed learning \cite{8865093}, and are mostly qualitative with a few exceptions~\cite{9357490,8714026,8865093}. Clearly, there is a lack of a holistic tutorial that provides a comprehensive overview on the field of distributed learning over wireless networks, while shedding light on ways to overcome the aforementioned challenges, for FL, MARL, and beyond.

\subsection{Contributions}
The main contribution of this article is a holistic tutorial that outlines the challenges and opportunities associated with distributed learning over wireless networks. We begin by focusing on FL which is perhaps the most renowned framework for distributed learning. In this context, we first provide a detailed overview on federated averaging, federated multi-task learning, and model agnostic meta learning based FL and summarize their drawbacks and usage. Then, we turn our attention on the possibility of performing joint learning and communications when FL is deployed over wireless networks. We first introduce four important performance metrics to quantify the FL performance over wireless networks and analyze how wireless factors affect these metrics. Next, we discuss novel approaches ranging from compression and sparsification, wireless resource management, FL training method design, and over-the-air computation (OAC or AirComp), to optimize the FL performance metrics, while taking into account the need for communication-efficient learning and effective distributed computing. For each approach, we introduce the motivation for optimizing the FL performance and then present a detailed literature review, an illustrative example, and future research opportunities. Next, we delve further into communication-efficient federated learning and present the concept of federated distillation and its ramificiations. We then discuss the use of distributed learning for inference and the use of MARL as a suitable framework for distributed learning and optimization. We conclude by providing industry insights on this emerging area.

The rest of this paper is organized as follows. In Section II, we introduce communication efficient FL. Section III presents AirComp based FL. In Section IV, we present federated distillation. Section V introduces distributed inference over wireless networks. In Section VI, we introduce multi-agent reinforcement learning over wireless networks. Finally, conclusions are drawn in Section VII.

\section{Communication Efficient Federated Learning}\label{se:FL}
Next, we first introduce the preliminaries of FL. In particular, we introduce the federated averaging and personalized FL algorithms. Then, we introduce four important performance metrics to quantify the FL performance over wireless networks and analyze how wireless factors  affect  these  metrics. We then present the research directions of deploying FL over wireless networks. Finally, open problems and industry interest of designing communication efficient FL are introduced.

\subsection{Preliminaries of FL}
Consider a set $\mathcal{U}$ of $U$ devices orchestrated by a parameter server (PS) to jointly train a common ML model. We assume that each participating device $i$ owns a dataset $\mathcal{K}_i$ of $K_i$ training samples, where each training sample $k \in \mathcal{K}_i$ consists of an input vector $\mathbf{x}_{i,k}$ and a corresponding output vector $\mathbf{y}_{i,k}$. Next, we introduce different FL problems. 

\subsubsection{Common Federated Learning}  
The training objective of common FL is given as follows:
\addtocounter{equation}{0}
\begin{equation}\label{eq:ML}
\begin{split}
\mathop {\min } \limits_{{\mathbf{m}}}  \sum\limits_{i = 1}^U \frac{p_i}{K_i}  \sum\limits_{k\in \mathcal{K}_i} {f\left( {{\mathbf{m}},{\mathbf{x}_{i,k}},{\mathbf{y}_{i,k}}} \right)},
\end{split}
\end{equation}
where $\mathbf{m} \in \mathcal{R}^d$ is the ML model that FL aims to find, $f\left(\cdot\right)$ is a loss function that captures the accuracy of the considered FL algorithm by building a relationship between an
input vector $\mathbf{x}_{i,k}$ and the corresponding output vector $\mathbf{y}_{i,k}$; $p_i$ is a scaling parameter that scales the weight of device $i$'s training loss, $\frac{1}{K_i}\sum\limits_{k\in \mathcal{K}_i} {f\left( {{\mathbf{m}},{\mathbf{x}_{i,k}},{\mathbf{y}_{i,k}}} \right)}$, on the total training loss with $\sum\limits_{i = 1}^U p_i=1$.  Problem (\ref{eq:ML}) is commonly solved by using iterative distributed optimization techniques, orchestrated by the PS.
Federated Averaging (FedAvg) \cite{mcmahan2017communication} is the first FL algorithm proposed by Google to solve problem (\ref{eq:ML}). The training process of FedAvg proceeds as follows:
 \begin{itemize}
\item[a.] The PS broadcasts the information to initialize the learning model to each device. 
\item[b.] Each device uses some local learning method, such as the stochastic gradient descent (SGD), to train its ML model (called local ML model).
\item[c.] Each device sends its updated ML model parameters to the PS.
\item[d.] The PS generates a common ML model $\mathbf{b}$ (called global ML model) using equation $\mathbf{b}\left(t+1\right)=\sum\limits_{i = 1}^U p_i \left(\mathbf{m}_{i}\left(t+1\right) - \mathbf{b}\left(t\right)\right)+\mathbf{b}\left(t\right)$, where $\mathbf{b}\left(t\right)$ is the global ML model at learning step $t$ and $\mathbf{m}_i\left(t+1\right)$ is a  $d$-dimensional parameter vector representing the local ML model of device $i$ at the end of learning step $t$.
\item[e.] The PS sends the generated common ML model back to all devices.
\item[f.] Steps from b. to e. are repeated for a fixed number of iterations, or some convergence criteria is met.    
\end{itemize} 
From the training procedure, we observe that, in FedAvg, each device transmits its gradient vector $\mathbf{m}_{i}\left(t+1\right) - \mathbf{b}\left(t\right)$ to the PS instead of sending its private data thus promoting data privacy for devices. Hereinafter, we define the implementation of steps from b. to e. as one \emph{learning step}. Meanwhile, at one step b., each device can update its ML model multiple times. Hereinafter, a device using the SGD method to update its ML model once is called one local update. Since FedAvg finds a common ML model for all devices, the training loss of each device will be significantly increased when the data distribution of each device is non independent and identically distributed (Non-IID).



To deal with Non-IID data, next, we introduce personalized FL. In particular, we introduce two classical personalized FL algorithms: federated multi-task learning \cite{DBLP:conf/nips/SmithCST17} and model agnostic meta learning (MAML) based FL \cite{NEURIPS2020_24389bfe}.

\subsubsection{Federated Multi-Task Learning} In federated multi-task learning (FMTL), devices are considered to implement correlated but different learning tasks. In other words, Non-IID data distributions of devices can be considered as different tasks. The training purpose of FMTL is given as follows:
\addtocounter{equation}{0}
\begin{equation}\label{eq:FMTL}
\begin{split}
\mathop {\min } \limits_{{\mathbf{M}},\mathbf{\Omega}}  \sum\limits_{i = 1}^U\sum\limits_{k\in \mathcal{K}_i} {f\left( {{\mathbf{m}_i},{\mathbf{x}_{i,k}},{\mathbf{y}_{i,k}}} \right)}+R\left(\mathbf{M},\mathbf{\Omega} \right),
\end{split}
\end{equation}
where $\mathbf{M}=\left[\mathbf{m}_1, \ldots, \mathbf{m}_U\right]$, $\mathbf{\Omega}$ models the relationship among different learning tasks of devices, and function $R\left( \cdot \right)$ is a regularizer. To solve problem (\ref{eq:FMTL}), one can use separate problem (\ref{eq:FMTL}) into several subproblems so as to enable devices to solve problem (\ref{eq:FMTL}) in a distributed manner. For example, the authors in \cite{DBLP:conf/nips/SmithCST17} used a dual method and quadratic approximation to divide problem (\ref{eq:FMTL}). Then, each device $i$ can individually optimize its ML model $\mathbf{m}_i$ under given $\mathbf{\Omega}$ while the PS updates $\mathbf{\Omega}$ using the updated $\mathbf{M}$. After the devices and the PS iteratively optimize $\mathbf{M}$ and $\mathbf{\Omega}$, problem (\ref{eq:FMTL}) can be solved.   

From (\ref{eq:ML}) and (\ref{eq:FMTL}), we can see that, in FedAvg, devices will have the same ML model at convergence. In contrast, in FMTL, devices may have different ML models at convergence. This is due to the fact that for Non-IID data or different learning tasks, devices with different ML models can achieve less sum training loss than devices with a common ML model.

\subsubsection{MAML Based FL}

MAML based FL aims to find a ML model using which each device can find a personalized ML model via one or a few steps of gradient descent. The training purpose of MAML based FL is given as follows:
\addtocounter{equation}{0}
\begin{equation}\label{eq:MAML}
\begin{split}
\mathop {\min } \limits_{{\mathbf{m}}}  \sum\limits_{i = 1}^U \frac{p_i}{K_i}  \sum\limits_{k\in \mathcal{K}_i} {f\left( {{\mathbf{m}-\lambda\nabla f_i},{\mathbf{x}_{i,k}},{\mathbf{y}_{i,k}}} \right)},
\end{split}
\end{equation}
where $\lambda$ is the learning rate and $\nabla f_i$ is the gradient descent of local ML model of device $i$. From (\ref{eq:MAML}), we can see that MAML based FL aims to find a common ML model for all devices. Then, the devices can use their own data to update their common ML models via a few steps of gradient descents so as to find the personalized ML models.

Given the overview of FedAvg, FMTL, and MAML based FL, we remark the following:

\begin{itemize}
\item FMTL directly optimizes the personalized ML model of each device while MAML based FL optimizes the initialization of ML model of each device. 
\item FedAvg is recommended for processing IID data while FMTL and MAML based FL are recommended for processing Non-IID data.
 \item Choosing between FMTL or MAML based FL depends on whether the PS knows the relationship among the data distributions of the devices.
  \item All FL algorithms must be trained by a distributed iterative process.
\end{itemize}

\subsection{Performance Metrics of FL over Wireless Netowrks}
Next, we introduce four key metrics that evaluate the performance of FL implemented over wireless networks: a) training loss, b) convergence time, c) energy consumption, and d) reliability.

\subsubsection{Training Loss} Training loss is the value of the loss functions $f\left(\cdot\right)$ defined in (\ref{eq:ML}), (\ref{eq:FMTL}), and (\ref{eq:MAML}). From the FL training procedure, we can see that the FL training loss depends on the ML models of all devices. In wireless networks, devices' ML models are transmitted over imperfect wireless links. Therefore, they may experience transmission errors thus negatively impacting the training loss. Meanwhile, due to limited energy and computational capacity, only a subset of devices can participate in FL. Therefore, only a subset of devices' ML models can be used to generate the global ML model thus negatively impacting the training loss.  

\subsubsection{Convergence Time} 
For FL implemented over wireless networks, its convergence time $T$ is expressed as 
\begin{equation}\label{eq:convergencetime}
T=\left(T_\textrm{C}+T_\textrm{T}\right)\times N_\textrm{T},
\end{equation} 
where $T_\textrm{C}$ is the time that each device used to update its local ML model at each learning step, $T_\textrm{T}$ is the maximum ML model transmission time per learning step, $N_\textrm{T}$ is the number of learning steps that FL needs to converge. From (\ref{eq:convergencetime}), we can see that FL convergence time depends on three components: a) ML parameter transmission delay $T_\textrm{T}$, b) the time $T_\textrm{T}$ needed by each device to train its local ML model, and c) number of learning steps $N_\textrm{T}$. Here, we need to note that $T_\textrm{C}$ and $N_\textrm{T}$ are dependent. In particular, increasing the number of SGD steps to update a local ML model at each learning step (e.g., increasing $T_\textrm{C}$) can decrease the number of learning steps $N_\textrm{T}$ that FL needs to converge.  

\subsubsection{Energy Consumption} The energy consumption $E$ of each device participating the entire FL training is expressed as
\begin{equation}\label{eq:energy}
E=\left(E_\textrm{C}+E_\textrm{T}\right)\times N_\textrm{T},
\end{equation}
where $E_\textrm{C}$ is energy consumption of each device training its ML model at each learning step and $E_\textrm{T}$ is the energy consumption of transmitting ML parameters to the PS at each learning step. From (\ref{eq:energy}), we can see that energy consumption of each device depends on three components: a) energy consumption for ML parameter transmission, b) energy consumption for training local ML model, and c) number of learning steps that FL needs to converge. Here, since increasing the number of SGD steps to update a local ML model at each learning step can decrease the number of learning steps $N_\textrm{T}$ that FL needs to converge, a trade-off exists between $E_\textrm{C}$ and $N_\textrm{T}$.

\subsubsection{Reliability} FL reliability is defined as the probability of FL achieving a target training loss.
For wireless FL, due to limited wireless resources, only a subset of devices can participate in the FL training at each learning step. Hence, the devices that transmit FL parameters to the PS at different learning steps may be different, which will affect the FL convergence time and training loss. Meanwhile, imperfect wireless links will cause errors on the FL parameters used to generate the global ML model, hence decreasing training loss. 

\begin{table*}[t!]\footnotesize
\setlength{\belowcaptionskip}{0pt}
\setlength{\abovedisplayskip}{3pt}
\newcommand{\tabincell}[2]{\begin{tabular}{@{}#1@{}}#2\end{tabular}}
 \setlength{\abovecaptionskip}{2pt}
  \caption{
    \vspace*{-0.2em}Summary of Effects of Communication Factors on FL Metrics}\label{table_1}\vspace*{0em}
\centering
\tabcolsep=0.11cm 
\scalebox{0.99}{
\begin{tabular}{|c|ccccccc|}
\hline
\multirow{2}{*}{\textbf{Communication Factors}}& \textbf{Training}& \textbf{Local Training}& \textbf{FL Parameter}& \textbf{Total Number of}& \textbf{Energy Used for}& \textbf{Energy Used for FL}&\multirow{2}{*}{\textbf{Reliability}} \\
&\textbf{Loss}&\textbf{Time $T_\textrm{C}$}& \textbf{Transmission Time $T_\textrm{T}$}&\textbf{Learning Steps $N_\textrm{T}$}&\textbf{Local Training $E_\textrm{C}$}&\textbf{Transmission $E_\textrm{T}$}&\\
\hline
Spectrum resource& $\surd$ & &$\surd$ &$\surd$&&$\surd$&$\surd$ \\
 \hline
Computational capacity & $\surd$& $\surd$& &$\surd$&$\surd$&& \\
 \hline
Transmit power & $\surd$ & &$\surd$&$\surd$  &&$\surd$&$\surd$\\
 \hline
Wireless channel & $\surd$ & &$\surd$&$\surd$&&$\surd$&$\surd$ \\
 \hline
Set of devices that &\multirow{2}{*}{$\surd$} &&\multirow{2}{*}{$\surd$} &\multirow{2}{*}{$\surd$}&&&\multirow{2}{*}{$\surd$} \\
participate in FL& & & &&&& \\
 \hline
Size of the FL parameters &\multirow{2}{*}{$\surd$} &\multirow{2}{*}{$\surd$} & &\multirow{2}{*}{$\surd$}&\multirow{2}{*}{$\surd$}&&\multirow{2}{*}{$\surd$} \\
trained by each device& & & &&&& \\
\hline
Size of the FL parameters & & &\multirow{2}{*}{$\surd$} &&&\multirow{2}{*}{$\surd$}&\\
transmitted by each device& & & &&&& \\
\hline
\end{tabular}
}
\vspace{-0.34cm}
\end{table*}

\subsection{Effects of Wireless Factors on FL Metrics}
Given the metrics defined in the previous subsection, we first explain how wireless network factors such as spectrum, transmit power, and computational capacity affect these FL metrics. Table \ref{table_1} summarizes the relationship between various wireless factors and FL performance metrics. In Table \ref{table_1}, a tick implies that the communication factor will affect the FL performance metrics. For example, the spectrum resource allocated to each device for FL parameter transmission will affect the training loss, FL parameter transmission time per learning step $T_\textrm{C}$, Energy consumption of FL parameter transmission $E_\textrm{C}$, and reliability of FL. Next, we explain how these wireless factors affect the FL performance metrics as follows:
\begin{itemize}
\item Spectrum resource allocated to each device determines the signal-to-interference-plus-noise ratio (SINR), data rate, and the probability that the transmitted FL parameters include errors. Hence, spectrum resource affects the training loss, $T_\textrm{T}$, $E_\textrm{T}$, and reliability.   
\item Computational capacity determines the number of SGD updates that each device can perform at each learning step. Hence, computational capacity affects the time and energy used for local training. Meanwhile, as the number of SGD updates decreases, the training loss increases and the number of learning steps that FL needs to converge increases. 
 \item Transmit power and wireless channel determine the SINR, data rate, and the probability that the transmitted FL parameters include errors. Therefore, as the transmit power of each device increases, the training loss, $T_\textrm{T}$, $N_\textrm{T}$, and reliability decrease but $E_\textrm{T}$ increases.
  \item In FL, as the number of devices that participate in FL increases, the training loss and $N_\textrm{T}$ decrease while $T_\textrm{T}$ and reliability increase.
  \item As the size of the FL parameters trained by each device increases, the FL training loss, reliability, and the total number of learning steps may decrease. However, the energy and time used for training FL model increases. 
\end{itemize}

\subsection{Research Directions of Deploying FL over Wireless Networks}
Next, we present a comprehensive overview on the key research directions that must be pursued for practically deploying FL over wireless networks. For each research direction, we
first outline the key challenges, and then we discuss the state of the art, while also providing a recent result.

\subsubsection{\textbf{Compression and Sparsification}}\label{ss:compression}

A major challenge in distributed learning, particularly over wireless channels, is the communication bottleneck due to the large size of the trained models. For emerging neural networks with hundreds of millions of training parameters, transmitting so many locally trained parameter values from each participating device to the PS at every iteration of the learning algorithm over a shared wireless channel is a significant challenge. 

We would like to note here that the transmission of locally trained model parameters to the PS over a noisy wireless channel is a joint source-channel coding problem. Indeed, considering the fact that the PS is interested in the average of the models, rather than the individual model updates from different devices, this can be classified as a joint source-channel function computation problem \cite{AirComp_b2, AirComp_b3}. In general, we do not have an optimal solution to such a problem, particularly in the practical finite blocklength regime. The conventional approach to this problem is to separate the compression of the neural network parameters from the transmission over the channel. This so-called `digital' approach converts all the local updates into bits, which are then transmitted over the channel as reliably as possible, and all the decoded `lossy' reconstructions are averaged by the PS. A more efficient method would be to directly map each locally trained model parameters to channel inputs in an `analog' fashion \cite{Amiri:TSP:20}. While we will explore this approach in Section \ref{se:OADL} in detail, here we focus on digital schemes, and assume that each device individually compresses its own parameters. 

Numerous communication efficient learning strategies have been proposed in the ML literature to reduce the amount of information; that is, the number of bits, exchanged between the devices and the PS per global iteration. 

We classify these approaches into two main groups; namely {\em sparsification} and {\em quantization}. We would like to highlight that, thanks to the separation between compression and transmission of compressed bits to the PS, these strategies are independent of the communication medium and the communication protocol employed to exchange model updates between the devices and the PS, as they mainly focus on reducing the size of the messages exchanged. Therefore,  these techniques can be incorporated into the resource allocation and device selection policies that will be presented below.

The objective of sparsification is to transform the $d$-dimensional model update $\mathbf{m}$ at a device to its sparse representation $\tilde{\mathbf{m}}$ by setting some of its elements to zero. Sparsification can also be considered as applying a $d$-dimensional mask vector $\mathbf{M}\in \left\{0,1\right\}^d$ on $\mathbf{m}$, such that $\tilde{\mathbf{m}} = \mathbf{M} \otimes \mathbf{m}$, where $\otimes$ denotes element-wise multiplication. We can define the \textit{sparsification level} of this mask by $\phi \triangleq  \vert\vert\mathbf{M}\vert\vert_{1} /d$, i.e., the ratio of its non-zero elements to its dimension. Note that, when conveying a sparse model update to the PS, rather than conveying the values of all $d$ values of the model update, each device needs to convey only the values of $\phi d$ non-zero values and their locations. Therefore, the lower the sparsification level, the higher the compression ratio, and the lower the communication load. It is known that when training a complex neural network model using stochastic gradient descent methods, model updates can be highly sparse. Indeed, it has been shown that when training some of the popular large-scale architectures, such as ResNet \cite{NN.DRL} or VGG \cite{NN.VGG}, sparsification levels of $\phi\in[0.01,0.001]$  provides significant reduction in the communication load with almost no loss in their generalization performance \cite{SGD.sparse0, SGD.sparse1}. 

Top-$K$ sparsification is probably the most common strategy used in distributed learning. In top-$K$ sparsification, each device constructs its own sparsification mask $\mathbf{M}_{i,t}$ at each iteration by identifying the $K$ values in its local update with the largest absolute values  \cite{SGD.sparse3, SGD.sparse4, SGD.sparse5}. A simpler alternative to top-$K$ is rand-$K$ sparsification \cite{SGD.sparse5}, which selects the sparsification mask $\mathbf{M}_{i,t}$ randomly from the set of masks with sparsification level $K$. Both rand-$K$ and top-$K$ are biased compression strategies. In the case of rand-$K$, unbiased model updates can be obtained by scaling $\mathbf{M}_{i,t}$ with $d/K$, albeit at the expense of increasing the variance, which is not desirable in practice \cite{SGD.sparse5}. Top-$K$ sparsification has been shown to outperform rand-$K$ in practical applications in terms of both the test accuracy and the convergence speed; however, top-$K$ sparsification requires sorting the elements of the model update vector at each iteration, which can significantly slow down the learning process. Moreover, as mentioned above, top-$K$ sparsification requires transmitting the location of the non-zero values within the model update vector, which increases its communication load, whereas this is not needed for rand-$K$ if a pseudo-random generator with a common seed is used across all the devices to generate the same mask. A time-correlated sparsification strategy is introduced in \cite{ozfatura:arXiv:20}, where a common mask is sent from the PS at each iteration to be employed by all the devices to remove the additional communication load due to sending locations of the non-zero values, and instead, each device sends only a limited number of significant values that are not present in this common mask, enabling exploration of more efficient masks. This approach exploits the time correlations between model updates across different iterations, and can provide up to 2000 times reduction in the communication load with minimal loss in model accuracy. We also note that, when employed for distributed training of DNN architectures, these sparse communication strategies can be applied to each layer of the network separately, since it is observed that different layers have different tolerance to sparsification of their weights \cite{SGD.sparse0, ozfatura:arXiv:20}. 

As mentioned above, the weights of a neural network take values from real numbers, and hence, even after sparsification they cannot be transmitted to the PS as they are, and must be quantized. In practice, since even the computing of local iterations are carried out using 32bit floating point representations, we can assume that each weight can be conveyed to the PS perfectly using 32 bits. Quantization techniques aim at identifying more efficient representations of the network weights that use less than 32 bits per weight \cite{SGD.q0, SGD.q1, TernGrad:17}. At the extreme, only a single bit can be used to represent only the sign of each element, which would result in a $32$ times reduction in the communication load. Sign based compression techniques for distributed optimization have been studied for a long time mainly to improve the robustness and convergence of learning algorithms \cite{RPROP:ICNN:93}. It has been recently shown that simple sign-based quantization together with majority voting converges to the optimal solution (under certain assumptions), and provides an extremely communication-efficient viable alternative in practice as well \cite{AirComp_f4, signSGD:19, SGD.q4}. A more advanced vector quantization scheme is considered in \cite{shlezinger:arXiv:20} and \cite{PNASchen}. 

We would like to highlight that,  most of the literature on distributed learning, and particularly its implementation over a wireless network, focus on the limitation of the uplink resources, and study quantization and sparsification of model updates from the devices while assuming that the global model from the PS is conveyed perfectly to all the participating devices. However, in the case of bandwidth-limited wireless networks, broadcasting the global model to all the wireless devices can be a challenge as well. The convergence of FL with noisy downlink transmission of the global model is studied in \cite{amiri2020convergence}, and both digital and analog transmission of global model updates is considered.

\subsubsection{{Wireless Resource Management}} 

As shown in Table \ref{table_1}, wireless resources such as spectrum, transmit power, and computational capabilities jointly determine the FL training loss, convergence time, energy consumption, and reliability. Due to limited resources in wireless networks, it is necessary to optimize resource allocation so as to enable wireless networks to efficiently complete the FL training process. However, analyzing the effects of resource allocation on the FL performance faces several challenges. First, FL training process is distributed and iterative, but it is challenging to quantify how each single model update affects the entire training process. Also, since each device only exchanges its gradient vector with the PS, the PS does not have any information about devices' local datasets, and cannot use sample distribution or the values of the data samples to decide how resource allocation will affect the FL convergence.

{State of the Art:} Now, we discuss a number of recent works on the optimization of spectrum resources for deploying FL over wireless networks. The authors in \cite{9264742,8737464,8664630} study the trade-off between the local ML model updates and global ML model aggregation so as to minimize the total energy consumption for local ML model training and transmission or the FL training loss. The authors in \cite{Amiri:TWC:21b, 9154285,9170917,9292468} study the use of gradient statistics to optimize the set of devices that participate in FL at each training round. The authors in \cite{8851249} assume that the local FL model transmitted by the device can be decoded by the PS only when the SINR is under the target threshold, and analyzed how user scheduling affects the FL convergence. In \cite{Abad:ICASSP20}, a hierarchical network architecture is considered, and it is shown that the global convergence can be accelerated if local training is enabled with the help of small base stations, which only occasionally communicate with the macro base station for global consensus. Local learning not only speeds up the learning process, but also reduces the energy consumption of communication due to short distance transmissions, and increases communication efficiency by frequency reuse across multiple small cells, enabling parallel local learning processes.

%

{Representative Result:} To build the relationship between local ML model transmission and SINR of each wireless link, in \cite{chen2019joint}, we assumed that the local ML parameters of each device is transmitted as a single packet. Hence, the imperfect wireless transmission may cause errors on the transmitted packets. Meanwhile, we assumed that the PS will directly abandon the erroneous local ML parameters and will not use them for the global ML model generation. To this end, we can use SINR to derive the probability (called packet error rate) of each local ML packet including errors caused by wireless transmission. Given this probability, one can analyze the expected FL convergence, as follows:
\begin{equation}\label{eq:FLconvergence}
{\color{black}
\begin{split}
&\mathbb E \left(F\left(\mathbf{b}_{t+1}\right)-F\left(\mathbf{b}^{*}\right)\right)  \leq A^t  \mathbb E\left(F \left(\mathbf{b}_{0} \right)-F\left(\mathbf{b}^{*}\right)\right) \\&\;\;\;\;\;\;\;\;
+\underbrace{\frac  {2\zeta_1}  {LK}\sum\limits_{i = 1}^U K_i  \left(1-a_i+a_iq_i\left(\mathbf{r}_{i}, P_{i} \right)\right) \frac{1-A^t}{1-A}}_{\textrm{Impact of wireless factors on FL convergence}},
 \end{split}
 }
\end{equation}
where $A<1$ is the convergence rate function of network parameters such as power, bandwidth, and link quality (see \cite[Theorem 1]{chen2019joint} for expression), $F\left(\mathbf{b}\right)= \sum\limits_{i = 1}^U \frac{p_i}{K_i}  \sum\limits_{k\in \mathcal{K}_i} {f\left( {{\mathbf{b}},{\mathbf{x}_{i,k}},{\mathbf{y}_{i,k}}} \right)}$, $a_i, \mathbf{r}_i, P_i$ are respectively the device selection index, resource allocation vector, and transmit power of device $i$, $q_i\left(\mathbf{r}_{i}, P_{i} \right)$ is the probability of the transmitted packet of device $i$ including errors, $\mathbf{b}_{t+1}$ is the global ML model at learning step $t+1$ while $\mathbf{b}^{*}$ the optimal global ML model that can solve problem (\ref{eq:ML}). From (\ref{eq:FLconvergence}), we can see that wireless factors (e.g., $a_i, \mathbf{r}_i, P_i$) and FL parameters (e.g., $K_i, U$,) jointly determine the FL convergence. From (\ref{eq:FLconvergence}), we can also see that as $t$ is large enough, $A^t  \mathbb E\left(F \left(\mathbf{b}_{0} \right)-F\left(\mathbf{b}^{*}\right)\right)=0$ but the second term will not be equal to 0. Therefore, we only need to minimize the second term via optimizing resource allocation and device selection.

Equation (\ref{eq:FLconvergence}) captures how the packet error rates and device selection affect the FL convergence. In a special scenario where all devices can participate in FL and all transmitted ML parameters are correct, the FL algorithm can find an optimal global ML model to solve problem (\ref{eq:ML}). According to this equation, one can analyze the effects of other wireless factors (e.g., device mobility, energy harvesting) that is related to packet error rates on the FL convergence.

\subsubsection{{FL Training Method Design}}
Beyond the use of wireless techniques, one can design novel FL training methods and adjust the learning parameters (e.g., step size) to enable FL to be efficiently implemented over wireless networks. 
Naturally, wireless devices have a limited amount of energy and computational resources for ML model training and transmission. In consequence, the size of ML model parameters that can be trained and transmitted by a wireless device is typically small and the time duration that the wireless devices can be used for training FL is  typically short. Hence, while designing FL training methods, the energy, computation, and training time constraints
need to be explicitly taken into account. Meanwhile, FL training methods determine the network topologies formed by the devices thus significantly affecting the FL training complexity and the FL convergence time. In consequence, designing FL training methods also needs to jointly consider the locations and mobility patterns of wireless devices as well as wireless channel conditions.  

{State of the Art:} Designing communication efficient FL training methods has been studied from various perspectives. In particular, an error feedback based SignSGD update method is proposed in \cite{SGD.q4} to improve both convergence and generalization. In \cite{Abad:ICASSP20}, hierarchical FL is proposed, where devices are grouped into clusters, and devices within each cluster carry out local learning with the help of a small base station or a cluster head, while a global model is trained at the macro base station. This framework is extended in \cite{hosseinalipour2020multistage}, which designed a training method for a multi-layer FL network. The authors in \cite{NIPS2018_7752} and \cite{NEURIPS2019_4e87337f} proposed a gradient aggregation method so as to decrease the number of devices that must transmit the local ML parameters to the PS thus reducing the FL communication overhead. In \cite{kassab2020federated}, the authors introduced a non-parametric generalized Bayesian inference framework for FL so as to reduce the number of learning steps that FL needs to converge. The authors in \cite{Lin2020Dont} proposed a post-local SGD update method that enables each device to update its ML parameters once in the initial multiple learning steps while update its ML parameters several times in the following learning steps. This post-local SGD method can significantly improve the generalization performance and communication efficiency. The work in \cite{yu2019parallel} designed a parallel restarted SGD method using which each device will average its ML model every certain learning steps and perform local SGDs to update its ML model in other learning steps.

{Representative Result:} To include more devices to participate in FL and reduce the devices' reliance on the PS, the authors in \cite{9311931, ozfatura2020decentralized}, and \cite{8340193} used decentralized averaging methods to update the local ML model of each device. In particular,  using the decentralized averaging methods, each device only needs to transmit its local ML parameters to its neighboring devices. Each device can use the ML parameters of its neighboring devices to estimate the global ML model. Therefore, using the decentralized averaging methods can reduce the communication overhead of FL parameter transmission. Meanwhile, since each device only needs to connect to its neighboring devices, the devices that cannot connect to the PS due to limited energy and wireless resources may be able to be associated with their neighboring devices so as to participate in FL training. Therefore, using the decentralized averaging methods for local ML model update can include more wireless devices to participate in FL training. Meanwhile, using decentralized averaging methods, the devices can form different network topologies to further improve FL parameter transmission time and ML model inaccuracy caused by imperfect wireless transmission. Finally, since each device shares its ML parameters with only its neighboring devices and the PS cannot know the ML parameters of all devices, privacy against the PS can be improved (assuming the neighbouring devices are trustworthy).               

 \begin{figure}[t]
  \centering
  {\subfigure[Simulation system]{\includegraphics[width=9cm]{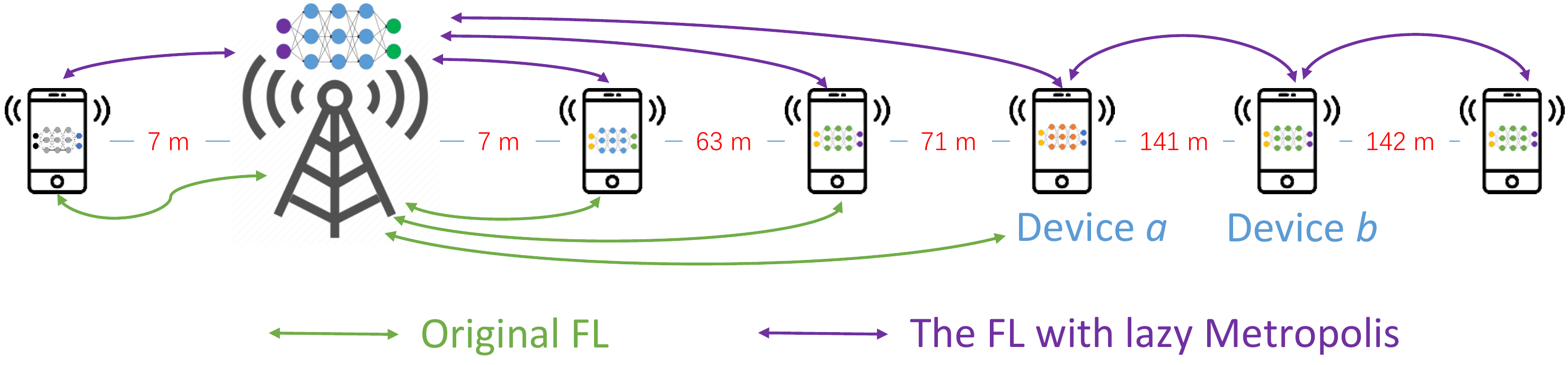}
\label{fig3a}}\hspace{0.1cm}
\subfigure[Simulation result]{\includegraphics[width=7cm]{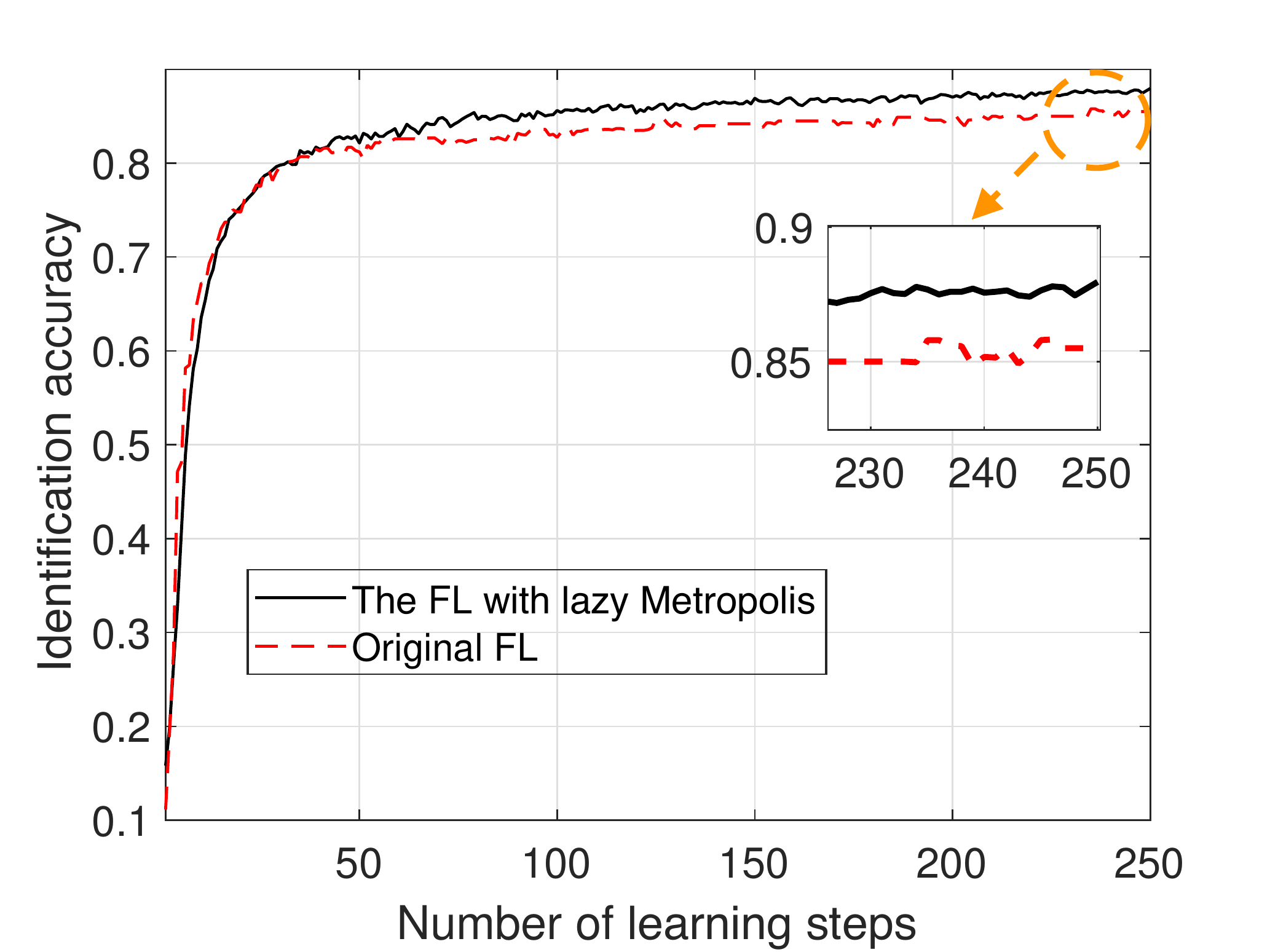}
\label{fig3b}} }
  \caption{Simulation system and result to show the performance of the FL with the lazy Metropolis update method. In this figure, a red digit is the distance between two adjacent devices.}\label{fig3}
\vspace{-0.3cm}
\end{figure}

To show the performance of the FL with the decentralized averaging method, particularly the lazy Metropolis update method (called the FL with lazy Metropolis hereinafter), we implemented a preliminary simulation for a network that consists of one base station that is acted as a PS and six devices, as shown in Fig. \ref{fig3a}. In Fig. \ref{fig3a}, the green and purple lines respectively represent the local ML parameter transmission of original FL and the FL with lazy Metropolis. Due to the transmission delay requirement, only 4 devices can participate in original FL. For the FL with the lazy Metropolis update method, 6 devices can participate in the FL training process since the devices that use the lazy Metropolis update method can connect to their neighboring devices. Therefore, from Fig. \ref{fig3b}, we can see that the FL with lazy Metropolis outperforms the original FL in terms of identification accuracy.
In fact, the FL with lazy Metropolis can also reduce the energy consumption for device $b$ since it only needs to transmit its ML model parameters to device $a$ instead of the BS.       

\subsection{Open Problems of Deploying FL over Wireless Networks}
Given the general research directions and challenges of deploying FL over wireless networks, next, we discuss open research
problems.

\subsubsection{Convergence Analysis}
FL convergence analysis results show the effects of wireless factors on key learning metrics; and hence, can be used to optimize the allocation of wireless resources and on deciding other wireless system parameters. 
For convergence analysis, there is a need to analyze how wireless factors affect the convergence of realistic FL with non-convex local ML models and loss function. Most existing works \cite{9264742,8737464,8664630,Amiri:TWC:21b, 9170917,9292468, 8851249, chen2019joint} use distributed optimization methods to analyze the effects of wireless factors on the FL convergence, assuming that the FL loss function is strongly convex and twice-continuously differentiable, and its gradient is uniformly Lipschitz continuous. However, realistic FL algorithms may not satisfy these conditions. Meanwhile, the convergence analysis should characterize the dynamics caused by SGD updates for local ML models, wireless channels, and device mobility. In addition, instead of finding the upper and lower convergence bounds in the existing works, the designed convergence analysis methods must find an exact convergence value and show the exact number of learning steps needed to converge.    

\subsubsection{Wireless Resource Management}
While there have been an increasing number of studies on the optimization of wireless resource allocation
for FL, there are still several many open problems, including: 1) considering the optimization of resource allocation based on the mobility patterns of devices, 2) jointly considering resource allocation, compression scheme design, and learning parameter (e.g. step size) adjustment so as to simultaneously reduce the time used for ML parameter training and transmission, 3) optimizing resource allocation for the devices that participate in FL, while guaranteeing the quality-of-service of other cellular-connected devices, and 4) adopting suitable frequency bands  (e.g., mmWave and Thz bands) for local and global ML parameter transmission.   

\subsubsection{Compression and Sparsification} For developing compression and sparsification schemes to improve FL performance metrics, there are several key problems. First, in wireless networks, link characteristics of each device will be different (e.g., different data rates). Hence, to efficiently use wireless resources for FL model transmission, it is necessary to design novel heterogeneous compression schemes that enable each device to encode its local FL model using different number of bits or different coding techniques. Second, since one can use gradient vectors to recover the raw data, called \textit{gradient leakage} \cite{NEURIPS2019_60a6c400}, it is necessary to design new compression or sparsification schemes that optimize FL performance metrics while considering data leakage. Although designing complicated compression or sparsification schemes can significantly reduce data leakage, it also introduces processing latency. Therefore, there is a need to design new compression or sparsification schemes that can significantly reduce data leakage while reducing FL convergence time.     

\subsubsection{FL Training Method Design}
Designing efficient FL training methods requires addressing a number of key problems. A fundamental problem is to enable the devices to form an optimal network topology that maximizes various FL performance metrics and trade-offs. This is a challenging problem since the solution must jointly account for the network topology, device heterogeneity, wireless dynamics, FL learning parameters (e.g., the data size of local ML model), and multiple dependent FL performance metrics. Other important open problems include: 1) designing asynchronous training methods while considering the network topology optimization, 2) designing FL  training methods for devices that may
not completely know the network architecture, other device locations, and network composition; and hence, can connect only to a limited number of devices, 3) designing mobility-aware FL training methods, and 4) designing FL training methods that optimize FL performance metrics over wireless links while preventing data leakage.

\subsection{Industry Interest} 

As we have already mentioned, centralized based algorithms cannot fulfil the low latency demands of near real time applications of 5G and beyond cellular networks, while at the same time satisfying security and privacy requirements. Therefore, approaches that keep local data on resource-constrained edge nodes (such as mobile phones, IoT devices or radio sites) and employ edge computation to learn a shared model for prediction have become increasingly attractive for the networking and IoT industry, and in recent years it have appeared several implementations of distributed ML.

In April 2017, Google published a blog post ~\cite{mcmahan2017federated} describing they had successfully tested an FL method with many Android mobile devices. Using a federated averaging algorithm, a global model had been trained and deployed on Android mobiles to suggest search queries based on typing context from Android Gboard. The mobile used the model stored on the device to predict search queries (such as suggesting next words and expressions) but training and model update would only take place once the mobile was connected to WiFi and charging. As such it was ensured that only the user has a copy of their data. 

Besides Google, many other industrial researchers have also recently started exploring FL. Intel ~\cite{sheller2018multi} used FL to do medical imagining where personal data used for training a global model is kept local. During MWC 2019 ByteLake and Lenovo~\cite{Lenovo2019} have demonstrated FL IoT industry application that enables IoT devices in 5G networks to learn from each other as well as makes it possible to leverage local ML models on IoT devices. 

As we discuss in this paper, despite the apparent opportunities FL offers in wireless networks it is still in its early stages, as there exist several critical challenges that need to be researched, especially for large scale telecom application, such as computational resource allocation for training FL models at edge devices, selection of users for FL, energy efficiency of FL implementation, spectrum resource allocation for FL parameter transmission, and design of communication-efficient FL. Nevertheless, the telecom industry has recently started industrially applying distributed ML to improve privacy when using ML for network optimization, time-series forecasting ~\cite{diaz2019federated}, predictive maintenance and quality of experience (QoE) modeling ~\cite{ickin2019privacy, Vandikas2019}. To better understand the potential of FL in a telecom environment, the Ericsson authors in~\cite{ Vandikas2019} have tested it on a number of use cases, migrating the models from conventional, centralized ML to FL, using the accuracy of the original model as a baseline. Their research has indicated that the usage of a simple neural network results in a significant reduction in network utilization, due to the sharp drop in the amount of data that needs to be shared. Besides being improved by 5G techniques, FL has also been integrated in the 5G Network Data Analytics (NWDA) architecture where it has been used to deal with 5G problems such as Network Data Analytics Function (NWDAF) ~\cite{isaksson2020secure},  in order to improve privacy. 


%

\section{FL with Over-the-Air Computation}\label{se:OADL} 

As discussed earlier, one challenge confronting the implementation of FL in wireless networks, called federated edge learning (FEEL), is to overcome the communication bottleneck, which arises from many devices uploading high-dimensional model updates (locally trained models or stochastic gradients) to a PS. Researchers have attempted to reduce the resultant communication  latency using different approaches such as excluding slow devices (``stragglers") \cite{AirComp_a1, AirComp_a2}, selecting only those devices whose updates can significantly accelerate learning \cite{AirComp_a3, NIPS2018_7752,Amiri:TWC:21a}, or compressing updates by exploiting their sparsity using the techniques outlined in Subsection \ref{ss:compression}. An alternative approach of our interest in this section is to design new multiple access schemes targeting FEEL. The main drawback of the classic orthogonal-access schemes (e.g., OFDMA or TDMA) is that they do not scale well with the number of devices. Specifically, the required radio resources increase linearly with the number of transmitters, or else the latency will grow linearly. A recently emerged approach, called over-the-air computation (OAC), which is also known as AirComp, can provide the needed scalability for multi-access in FEEL. Fundamental limits of OAC have been studied in \cite{Nazer:IT:07, Soundararajan:IT:12} from an information theoretic perspective. A similar idea was considered in \cite{Mergen:TSP:06} for the distributed estimation of a discrete variable. OAC is applied to wireless communications in \cite{Goldenbaum:TC:13, OAC:Chen:18,  AirComp_f1}. Specifically, the deployment of AirComp to support FEEL, termed AirComp-FEEL, exploits the wave-form superposition property of a multi-access channel together with simultaneous transmission to realize over-the-air model/gradient aggregation. Given simultaneous access, the latency becomes independent of the number of devices. This overcomes the communication bottleneck to facilitate the implementation of FEEL over many devices. In this section, we shall first discuss the basic principle and techniques of AirComp, and then explore its deployment in a communication-efficient FEEL system. 

\subsection{AirComp Principle and Techniques} 
\subsubsection{AirComp Principle} The mentioned idea of AirComp is elaborated as follows.  Given simultaneous time-synchronized transmission by devices, their signals are superimposed over-the-air and their weighted sum, called the aggregated signal, is received by the PS, where the weights correspond to the channel coefficients. For AirComp-FEEL, it is desirable to have uniform weights so that the aggregated signal is not biased towards any device, and can be easily converted to the desired average of the transmitted signals (i.e., model updates). To make this possible requires each device to modulate its signal using linear analog modulation and to invert its fading channel by transmission-power control. The former operation is necessary to exploit the channel's analog-waveform superposition property and the latter aligns the received magnitudes of individual signal components, called magnitude alignment. One may question the optimality of the use of seemingly primitive analog modulation compared with sophisticated digital modulation and coding. Interestingly, from the information-theoretic perspective, it was shown in \cite{AirComp_b2} that AirComp can be optimal in terms of minimizing the mean squared error (MSE) distortion if all the multi-access channels and sources are Gaussian and independent.

\begin{figure*}[t] 
	\centering
	\includegraphics[width=0.7\linewidth]{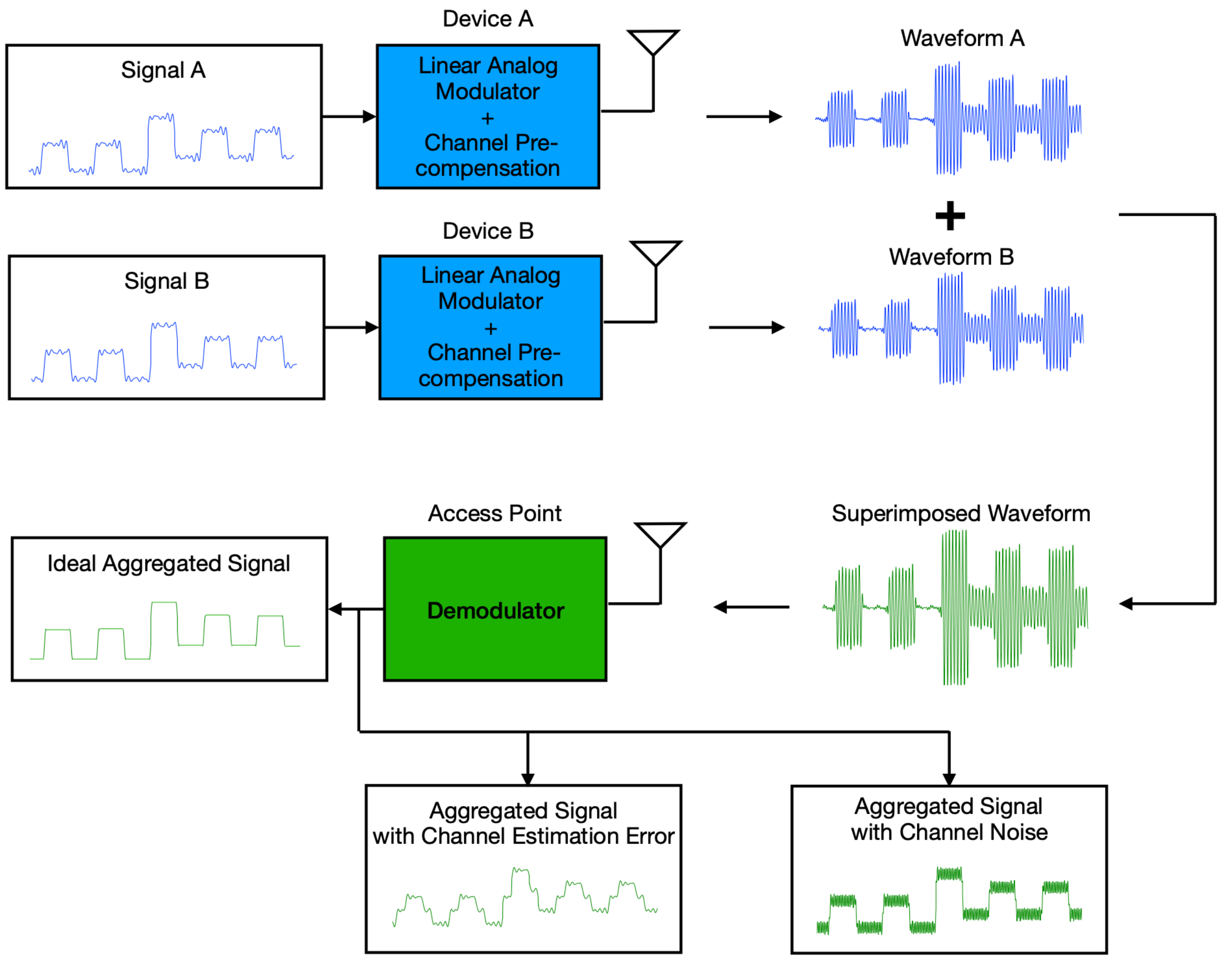}
	\caption{The principle of  AirComp \cite{AirComp_f1}.} \label{Fig_Performance}
\end{figure*}


Though FEEL requires only over-the-air averaging, AirComp is capable of computing a broad class of so called nomographic functions \cite{Goldenbaum:TC:13, AirComp_b3}. They are characterized by a post-processing function of a summation form with each term being a pre-processing function of an individual data sample. Besides averaging, other examples include arithmetic mean, weighted sum, geometric mean, polynomial, and Euclidean norm. Consequently, except for averaging, the implementation of AirComp of a nomographic function usually requires pre-processing of data before transmission and post-processing at the receiver. For a general function, it can be decomposed as a summation form of nomographic functions  \cite{AirComp_b4}. This suggests the possibility of approximately computing a general function with AirComp.

A key requirement for implementing AirComp is time synchronization of devices' transmissions. Such requirements also exist for uplink transmission (e.g., TDMA and SC-FDMA) in practical systems (e.g., LTE and 5G). In such systems, a key synchronization mechanism is called ``timing advance'', which can be also adopted for AirComp synchronization. The  technique of timing advance involves each device estimating the corresponding propagation delay and then transmitting in advance to ``cancel'' the delay. Thereby, different signals can arrive at the base station in their assigned slots (in the case of TDMA) or overlap with sufficiently small misalignment (in the cases of SC-FDMA and AirComp). Considering a synchronization channel for the purpose of propagation-delay estimation, its accuracy is proportional to the channel bandwidth \cite{AirComp_b5}. For instance, the estimation error is no larger than 0.1 microsecond for a bandwidth of 1 MHz. If AirComp is deployed in a broadband OFDM system (see the next sub-section), the error gives rise to only a phase shift to a symbol received over a sub-channel so long as the error is shorter than the cyclic prefix (CP). Then the phase shift can be compensated by sub-channel equalization. In an LTE system, the CP length is several microseconds, and hence more than sufficient for coping with synchronization errors.  This suggests the feasibility of AirComp deployment in practical systems. The impact of potential remaining synchronization errors on the performance of AirComp and techniques to tackle them have been recently studied in \cite{shao2021federated}. 

The distortion of digital modulation originates from quantization and decoding errors. In contrast, for AirComp, the main source of signal distortion is channel noise and interference that directly perturb analog modulated signals. Hence, a commonly used performance metric for AirComp is the MSE distortion of received functional values with respect to the ground-truth. In the context of AirComp-FEEL, channel noise and interference perturb the model updates and their effects can be evaluated using the relevant metric of learning performance. Finally, it is worth mentioning that AirComp is similar to non-orthogonal multiple access (NOMA) in both being simultaneous-access schemes. However, the distinction of AirComp is the harnessing of ``inference'' for functional computation via devices' cooperation. On the other hand, NOMA attempts to suppress inference as the devices (subscribers) transmitting independent data compete for the use of radio resources. 

\subsubsection{Broadband AirComp} 

In a practical broadband system, the spectrum is divided into sub-carriers using the OFDM modulation. The deployment of AirComp-FEEL in such a system involves simultaneous over-the-air aggregation of model-update coefficients transmitted over sub-carriers subject to power constraints of individual devices. The channel inversion discussed in the preceding sub-section need be generalized to the case of multiple sub-carriers as follows. Consider a specific uploading device. For each OFDM symbol, ideally each sub-carrier is linearly analog modulated with a single model/gradient element, whose power is determined by channel inversion. However, due to the power constraint, it is impractical to invert those sub-carriers in deep fade; hence they are excluded from transmission, called \textit{channel truncation}. AirComp requires all devices to have fixed and identical mappings between update coefficients to sub-carriers. As a result, channel truncation results in the erasure of coefficients mapped to sub-carriers in deep fade as they cannot be remapped to other sub-carriers. Channel truncation can potentially have a near-far problem where the fraction of erased coefficients, called truncation ratio, is much larger for a nearer device from the PS (hence with larger severe path loss) than a faraway device (will smaller loss). The problem introduces bias and degrades the learning performance. One solution is to apply channel truncation based only on small-scale fading with two-fold advantages: 1) approximately equalizing truncation ratios among devices, and 2) allowing the PS to exploit data even at faraway devices. The resultant scheme of truncated channel inversion scales the symbol transmitted over the $m$-th sub-carrier by a coefficient $p_k^{(m)}$ given as: 
\begin{equation}\label{Eq:Chan_Inverse}
p_k^{(m)} = \left\{
\begin{aligned}
&\frac{\eta}{r_k^{-\frac{\alpha}{2}}h_k^{(m)}}, && |h_k^{(m)}|^2 \geq g_{\text{th}}\\
&0, && \text{otherwise}
\end{aligned}
\right.    
\end{equation}
where $r_k^{-\frac{\alpha}{2}}$ is the path loss and   $h_k^{(m)}$ the fading gain. The parameter $\eta$ represents the aligned received magnitude of different signal components and is chosen by observing individual power constraints of all devices.  Next, given truncated channel inversion, the PS demodulates a certain number of OFDM symbols and thereby receives from the sub-carriers an over-the-air aggregated model update. This is then used to update the global model.

\subsubsection{MIMO AirComp }

MIMO (or multi-antenna) communication is widely adopted in practical systems (e.g., LTE and 5G) to support high-rate access by spatial multiplexing of data streams. The deployment of AirComp-FEEL in a MIMO system can leverage spatial multiplexing to reduce the communication latency by a factor equal to the multiplexing gain. Realizing the benefit requires the design MIMO AirComp, a technique multiplexing parallel over-the-air aggregation or equivalently AirComp of vector symbols, each comprising multiple update coefficients. The main distinction of MIMO AirComp is the use of receive beamforming, called aggregation beamforming, to enhance the received signal-to-noise ratios (SNRs) of aggregated observations from the PS array. The intuition behind the design of aggregation beamforming is that in terms of subspace distance, the beamformer should be steered aways from the relatively strong MIMO link and closer to those relatively weaker links. The purpose is to enhance the received SNRs of the latter at the cost of those of the former, thereby equalizating their channel gains. This facilitates the subsequent spatial magnitude alignment to enhance post-aggregation SNRs. Given the aggregation beamformer, the effective MIMO channel can be inverted at each device to implement spatial magnitude alignment after aggregation beamforming. Finding the optimal aggregation beamformer is a non-convex program and intractable. An approximate solution can be found in closed form though to mathematically express the above design intuition. Specifically, the received SNRs of spatial data streams from an individual MIMO link as observed after aggregation beamforming can be approximated using the smallest SNR, corresponding to the weakest eigenmode of the effective channel. Using the approximation, an approximate of the optimal aggregation beamformer can be obtained as the first $L$ left eigenvectors of the following matrix: $\mathbf{G} = \lambda^2_{\min, k} \mathbf{U}_k\mathbf{U}_k^H$ where $\lambda^2_{\min, k}$ is the smallest singular value of the $k$-th link and $\mathbf{U}_k$ its left $L\times 1$ eigen subspace \cite{AirComp_b3}. The matrix suggests that the aggregation beamformer is a weighted centroid of the eigen subspaces of individual MIMO links, where the weights are their smallest eigenvalues. This is aligned with the design intuition mentioned above.

\subsection{Design of AirComp Federated Learning}

\begin{figure*}[t] 
	\centering
	\includegraphics[width=0.6\linewidth]{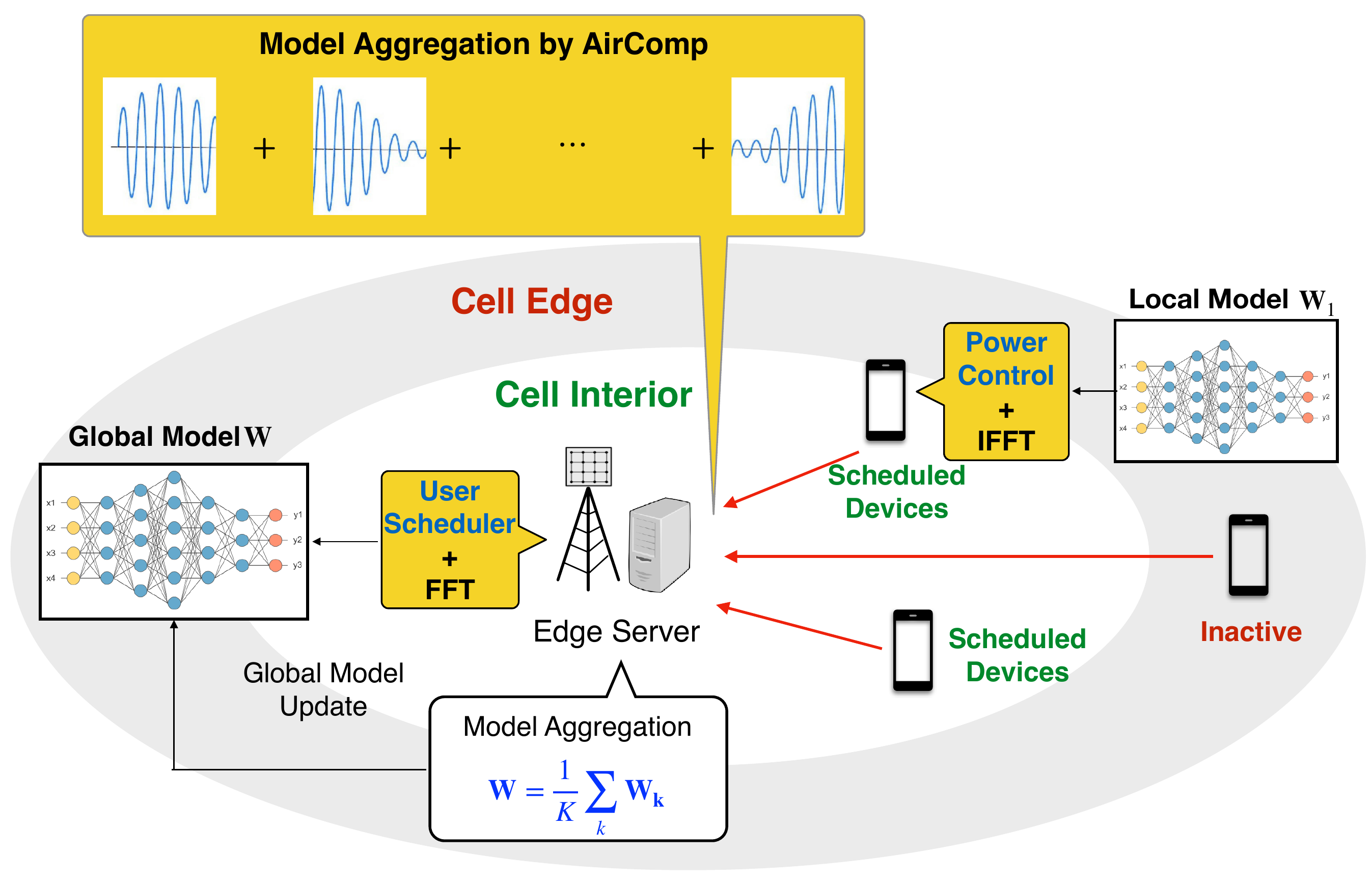}
	\caption{The AirComp-FEEL system.} \label{Fig_AirComp_FEEL_Sys}
\end{figure*}

Consider the  AirComp-FEEL system  in Fig.~\ref{Fig_AirComp_FEEL_Sys}. In this section, we discuss several issues concerning the design of such a system. 

\subsubsection{Model update distortion}

Given the deployment of broadband AirComp, the received aggregated model update at the PS is distorted in two ways. First, the local-model update transmitted by each device may lose some coefficients due to truncated channel inversion in \eqref{Eq:Chan_Inverse}. Second, the uncoded aggregated update is directly perturbed by channel noise. There exist a trade-off between these two factors. The sub-carrier/coefficient truncation ratio can be reduced by lowering the truncation threshold in \eqref{Eq:Chan_Inverse}. As a result, sub-carriers with small gains are used for transmission, and thus involved in channel inversion, consuming more transmission power. Due to individual devices' fixed power budgets, the magnitude alignment factor, $\eta$ in \eqref{Eq:Chan_Inverse}, has to be reduced. This leads to reduction on the received SNR and more noisy aggregated update received by the PS. For this reason, there exists a trade-off between the truncation ratio of each local-model update and the received SNR. In system design, such a trade-off should be balanced so as to regulate the overall distortion to prevent it from significantly degrading the learning performance. The operating point on this trade-off may also be adjusted along the iterations of the learning process as more accurate estimates of the model updates are needed as the learnign process gradually converges to its optimal value. 

\subsubsection{Device Scheduling} 

In a conventional radio-access system, its throughput or link reliability can be enhanced by scheduling cell-interior users at the cost of quality-of-service of cell-edge users. In the context of AirComp FEEL, the penalty of doing so is some loss of data diversity since the data at cell-edge devices cannot be exploited for model training, which can significantly reduce the generalization power of the learned model. To elaborate, due to the required signal-magnitude alignment in AirComp, the received SNR of aggregated model update is dominated by the weakest link among the participating devices. Consequently, including faraway devices with severe path loss can expose model updates to strong noise, and hence potentially slow down convergence and reduce model accuracy. On the other hand, including more devices, which are data sources, means more training data; from this perspective, they may have the opposite effects from the above. Therefore, designing a scheduling scheme for AirComp FEEL needs to balance this trade-off between update quality and data quantity. For example, when the device density is high, the path-loss threshold for selecting contributing devices can be raised and vice versa. On the other hand, mobility can alleviate this issue even when only cell-interior devices are employed. They are mobile and hence change over rounds, which benefits model training by providing data diversity. In the scenario with low mobility, one can also alleviate the issue by alternating cell-edge and cell-interior devices over different rounds \cite{AirComp_f2}. 

\subsubsection{Coding Against Interference}
Existing AirComp with uncoded linear analog modulation exposes model training to interference and potential attacks. Most existing works target single-cell systems and overcomes the noise effect by increasing the transmission power. However, in the scenarios of multi-cell networks or multiple coexisting services, the signal-to-interference ratios are independent of power. Besides coping with interference, making FEEL secure is equally important. This motivates the need of coding in AirComp. Possible methods include scrambling signals using pseudo-random spreading codes from spread spectrum or encoding the signals using Shannon-Kotelnikov mappings from joint source-channel coding \cite{AirComp_f5}, prior to their transmission. Both coding schemes have the potential of providing the desired property that AirComp remains feasible after coding so long as participating devices apply an identical code (spreading code or Shannon-Kotelnikov mapping) while interference is suppressed by despreading/decoding at the base station.

\subsubsection{Power Control}
Channel inversion is adopted in typical AirComp to realize magnitude alignment. Its drawbacks are to either exclude devices with weak links from FEEL at the cost of data diversity or consume too much power by inverting such links. In other words, channel-inversion transmission is sub-optimal in terms of minimizing the errors in aggregated gradients/models. Targeting a sensor system with i.i.d. data sources, it was shown in \cite{AirComp_f6} that the optimal power-control policy for error minimization exhibits a threshold based structure: when its channel gain is below a fixed threshold, a device should transmit with full power; otherwise it should adopt  channel inversion.  Nevertheless, the assumption of i.i.d. data sources does not hold for AirComp FEEL since stochastic gradients or local models of different devices are highly correlated. It is proposed in \cite{AirComp_f3} that information on gradient distribution can be exploited in power control for AirComp FEEL. While this provides significant gains in learning accuracy, the optimal power-control strategy in general remains an open problem.

\subsection{Performance  of AirComp Federated Learning}
First, let us consider the performance of AirComp FEEL in terms of model convergence. In the literature of FL, the convergence of the SGD algorithm is usually analyzed by making several typical assumptions on the loss function so as to allow tractability. First, the function is assumed to have a minimum value. Second, the gradients of the function are assumed to be smooth (e.g., Lipschitz continuous). Third, a  stochastic gradient computed at each device is assumed to be an unbiased estimate of the gradient of the function with bounded variance. For some special SGD algorithms such as ``signSGD" with one-bit quantization of gradient coefficients, additional assumptions are needed, such as a symmetric uni-modal distribution of the stochastic gradients \cite{AirComp_f4}. Under such assumptions, the existing methods of convergence analysis can be extended to AirComp FEEL. Essentially, it requires modification of a method to account for attenuation of aggregated gradients by channel fading and perturbation by channel noise.

As a concrete example, let us consider the implementation of ``signSGD'' in a broadband system supporting  AirComp FEEL. This requires the replacement of analog linear modulation with binary modulation (BPSK). In this case, the decision at the PS depends on the sign of the received signal, and corresponds to an ```over-the-air majority voting'' scheme that converts the received aggregated gradient into a binary vector \cite{AirComp_f4}. Given a general loss function, a common metric measuring the level of model convergence is the averaged (aggregated) gradient norm over rounds, denoted as $\bar{G}$. The  expectation of $\bar{G}$ for the considered system can be analyzed as a function of given numbers of rounds and devices, which quantifies the convergence speed. Specifically, it is shown in \cite{AirComp_f4} that 
\begin{equation}\label{Eq:Convergence}
\mathsf{E}\left[\bar{G}\right] \leq \frac{a}{\sqrt{N}}\left(f_1 + \frac{1}{\sqrt{K}}f_2 + b\right),
\end{equation}
where the factors $f_1$ and $f_2/K$ correspond to the descent using ground-truth gradients and the expected deviation of an aggregated  gradient  from its ground truth. The two parameters $a$ and $b$ capture the effects of wireless channels. In the ideal case with perfect channels, the parameters take on the values of $a = 1$ and $b = 0$. If the channels are AWGN, they are given as follows: 
\begin{equation}
    a_{\text{AWGN}} = \frac{1}{1 - \frac{1}{K\sqrt{\text{SNR}}}}, \quad b_{\text{AWGN}} = \frac{f_2}{K\sqrt{\text{SNR}}}\nonumber
\end{equation}
where the SNR refers to transmit SNR of a device. One can observe that they converge to their ideal-channel counterparts as the factor $K\sqrt{\text{SNR}}$ grows, where $K$ suppresses noise by aggregation and $text{SNR}$ by increasing signal power. If the channel has fading, then a transmitted local gradient can be truncated as we discussed. Let $\alpha$ denote the probability that a sub-carrier is truncated. Then, the two parameters in this case are given as
\begin{equation}
    a_{\text{FAD}} = \frac{1}{1 - (1 - \alpha)^K - \frac{2}{\alpha  K\sqrt{\text{SNR}_{\text{av}}}}}, \quad b_{\text{FAD}} = \frac{2f_2}{\alpha K\sqrt{\text{SNR}_{\text{av}}}}. \nonumber
\end{equation}
One can see that fading slows down the convergence rate with respect to the AWGN channel since $a_{\text{FAD}} > a_{\text{AWGN}}$ and $b_{\text{FAD}} > b_{\text{AWGN}}$. If the truncation probability $\alpha = 0$, the speeds for both cases are equal since fading is not severe, or the transmission power is sufficiently large to counteract it. 

The analysis  of convergence speed is useful for estimating the required number of communication rounds for model training. For FEEL, an alternative and perhaps more practical performance metric that can account for multi-access latency is the learning latency (in seconds). It accumulates per-round latency over the total rounds, which is determined by the convergence analysis. On one hand, AirComp achieves a lower model accuracy  than the conventional digital orthogonal access due to lack of coding. On the other hand, when there are many devices, AirComp dramatically reduces multi-access latency with respect to the latter. To have an idea on their relative performance, some experimental results from \cite{AirComp_f2} are shown in Fig.~\ref{Fig_Performance}. The experiment simulates the AirComp-FEEL system in Fig.~\ref{Fig_AirComp_FEEL_Sys} over broadband channels with  $100$ edge devices. The task is to train a convolutional neural network using  the distributed MNIST data for handwritten digit recognition. The update aggregation is performed by AirComp or OFDMA with adaptive modulation over a  broadband channel consisting of $1000$ orthogonal sub-channels. FEEL is implemented with local-model uploading.  For OFDMA, local-model parameters are  quantized using a $16$-bit scalar quantizer. The bit sequences of a local model are modulated onto sub-carriers using the classic  scheme of adaptive QAM modulation targeting a target bit-error-rate of $10^{-3}$. The averge received SNR is set as $10$ dB. One can observe from Fig.~\ref{Fig_Performance} that under such settings, compared with digital transmission, AirComp can reduce the communication latency by a factor approximately equal to the number of devices without significant loss of the learning accuracy.

\begin{figure*}[t] 
	\centering
	\includegraphics[width=0.8\linewidth]{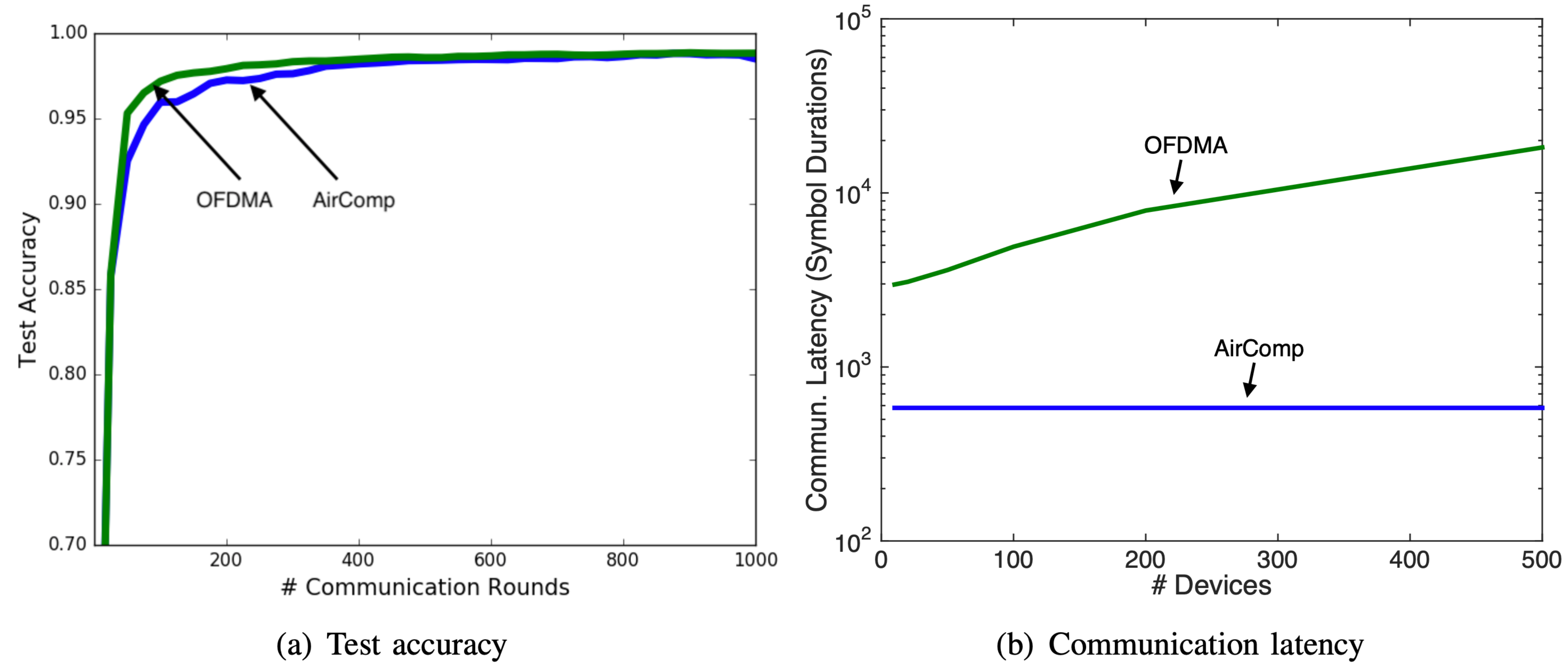}
	\caption{Learning accuracy and latency comparisons between FEEL with AirComp and digital OFDMA.} \label{Fig_Performance}
\end{figure*}

\subsection{State-of-the-Art and Research Opportunities }

The next-generation IoT is expected to connect tens of billions of edge devices and bring to them computing capabilities and intelligence. Among many others, a specific class of IoT applications has emerged, which requires an edge PS (which can be a base station) to aggregate data distributed at devices with wireless connectivity, termed wireless data aggregation (WDA). Such applications include distributed learning, the topic of this article, as well as vehicle platooning, drone swarm control, and distributed sensing. In such applications, a PS is interested in computing a function of distributed data generated by devices. The applications are either data intensive (e.g., distributed learning) or latency critical (e.g., vehicle platooning). The requirements have motivated researchers to develop AirComp to enable efficient WDA over many devices \cite{AirComp_f1}. In the area of FEEL, researchers overcome the communication bottleneck by applying AirComp to implement over-the-air aggregation of local updates. 

\textit{Analog compression.} Another challenge in AirComp is to enable learning in a broadband communication scenario. When the model updates are transmitted in an uncoded fashion, each iteration of the learning process requires as many channel resources as the model dimension, which can be very large in modern deep learning architectures. For example, architectures used for machine vision applications typically have millions of parameters; well-known AlexNet, ResNet50, and VGG16 architectures have 50, 26, and 138 million parameters, respectively. Models for natural language processing applications typically have much larger networks, with even billions of parameters. Therefore, training such large networks with uncoded transmission would require extremely large bandwidth, often not available at the network edge. This challenge is overcome in \cite{Amiri:TSP:20, amiri:TWC:20} by exploiting the sparsity of the model updates. However, note that, sparsification in the case of digital communication of the model updates requires the additional transmission of the index information of the transmitted model parameters from each device, adding a significant additional communication load. In \cite{Amiri:TSP:20, amiri:TWC:20}, the authors employ random projection of the sparsified model updates at the devices, which allows the devices to significantly reduce the bandwidth requirement without sacrificing the performance. 

As we discussed previously, researchers have also designed AirComp FEEL systems over multiple-antenna channels \cite{AirComp_e18, AirComp_b3}. While the beamforming vectors are optimized in \cite{AirComp_b3} to exploit the available multiple antennas for FL, it is shown in \cite{Amiri:TWC:21a} that if there are sufficiently many receive antennas at the PS, this can compensate for the lack of channel state information at the transmitter. It is further shown in \cite{Amiri:TWC:21a} that, since only the summation of the transmitted symbols needs to be decoded at the receiver, this also reduces the channel state estimation requirements at the receivers, which only needs an estimate of the sum channel gain from the devices to each antenna. An energy-aware device scheduling algorithm is studied in \cite{Sun:ICC:20} in conjunction with AirComp, where data-redundancy across devices is exploited to tackle with non-iid data distribution. 

\textit{Privacy in FL with AirComp.} As it was highlighted in \cite{amiri:TWC:20}, another important potential benefit of AirComp in the FL setting is regarding privacy. Even though FL has been proposed as a privacy-sensitive learning paradigm as the devices only transmit their model updates to the PS and the datasets remain localized, it has been shown that the gradient information can reveal significant information about the datasets, called \textit{gradient leakage} \cite{NEURIPS2019_60a6c400, Melis:SP:19}. Several works have proposed privacy mechanisms to prevent gradient leakage. In particular, differential privacy (DP) is used as a rigorous privacy measure in this context \cite{dwork2014algorithmic}. A common method to provide DP guarantees is to add noise to data before sharing it with third parties. In the digital implementation of FL, each device can add noise to its local gradient estimate before sharing it with the PS \cite{abadi2016deep, malekzadeh2021dopamine}, which results in a trade-off between privacy and the accuracy of learning. However, note that, the gradients (or, model updates) in the case of AirComp are received at the PS with additional channel noise. Several recent works have developed privacy-aware AirComp schemes based on this observation. In \cite{seif2020}, if the channel noise is not sufficient to satisfy the DP target, some of the devices transmit additional noise, benefiting all the devices. Instead, in \cite{Koda2020DifferentiallyPA} and \cite{Liu:JSAC:21}, transmit power is adjusted for the same privacy guarantee. While these works benefit mainly from the presence of channel noise, and depend critically on the perfect channel knowledge at the transmitters, in \cite{hasircioglu2021private}, the authors exploit the anonymity provided by AirComp for privacy, which prevents the PS to detect which devices are participating in each round.

AirComp FEEL is still in its nascent stage. There exist many promising research opportunities. For example,  AirComp FEEL can be wirelessly powered to lengthen devices' short battery lives due to intensive computation. As another example, efficient channel feedback based on the AirComp principle can be designed to suppress the excessive feedback overhead where devices are many \cite{AirComp_b3}. Furthermore, the deployment of AirComp FEEL in a multi-cell system exposes learning performance to the effect of inter-cell interference. Quantifying the effect can provide useful guidelines for network designers.

\section{Federated Distillation}\label{se:FD}

 Although federated learning is communication-efficient by nature, it still requires the exchange of large models over the air. Indeed modern deep neural network (NN) architectures often have a large number of model parameters. For instance, GPT-3 model is a state-of-the-art NN architecture for natural language processing (NLP) tasks, and has $175$ billion parameters corresponding to over $350$\,GB \cite{brown2020language}. Exchanging the sheer amount of deep NN model parameters is costly, hindering frequent communications particularly under limited wireless resources. Alternatively, federated distillation (FD) only exchanges the models' outputs whose dimensions are much smaller than the model sizes (e.g., 10 classes in the MNIST dataset). For instance, in a classification task, each device runs local iterations while storing the average model output (i.e., logit) per class. Then at a regular interval, these local average outputs are uploaded to the PS aggregating and averaging the local average output across devices per class. Subsequently, the resultant global average outputs are downloaded by each device. Finally, to transfer the downloaded global knowledge into local models, each device runs local iterations with its own loss function in addition to a regularizer measuring the gap between its own prediction output of a training sample and the global average output for the given class of the sample. Such  regularization method is called knowledge distillation (KD). 

 \begin{figure}
	\centering
	\includegraphics[width=9cm]{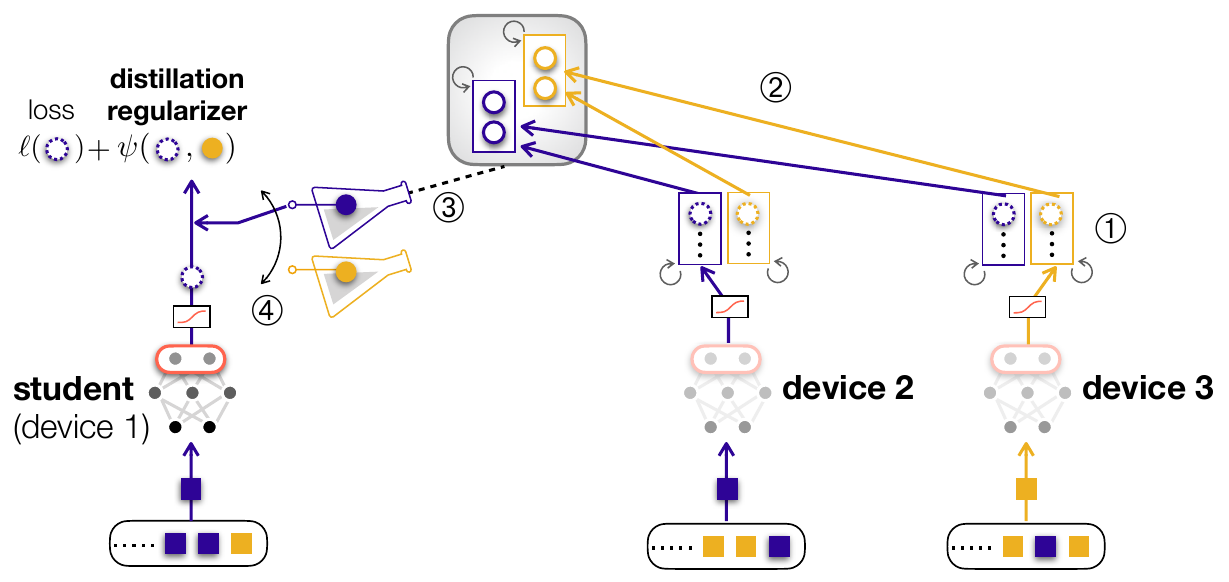}
	\caption{\small{A schematic illustration of federated distillation (FD) with 3 devices and 2 labels in a classification task.}} \label{Fig:Overview_FD}
	\end{figure}

\subsection{State-of-the-Art}  While FD was proposed in \cite{MLPCD}, its effectiveness is not limited to simple classification tasks under a perfectly controlled environment. In \cite{Han:Intellisys20}, FD is extended to an RL application by replacing the aforementioned pre-class averaging step of FD with an averaging operations across neighboring states for an RL task. In \cite{Ahn:ICASSP20} and \cite{Ahn:PIMRC20}, FD is implemented in a wireless fading channel, demonstrating comparable accuracy under channel fluctuations and outages with much less payload sizes compared to FL. In \cite{MixFLD}, a new technique called mix2FLD  was proposed   whereby local model outputs are uploaded to a PS in the uplink  whereas  global  model  parameters are  downloaded  in  the  downlink  as  in FL.
To  preserve  privacy  while  not  compromising  accuracy, linearly mixed-up  local  samples  are  uploaded,  and  inversely  mixed  up across different devices at the PS. 
For a comprehensive survey on the topic, please refer to \cite{seo2020federated, Google:FL19}

\begin{figure}[t] 
	\centering
	\subfigure[Test accuracy.]{\includegraphics[width=.85\linewidth]{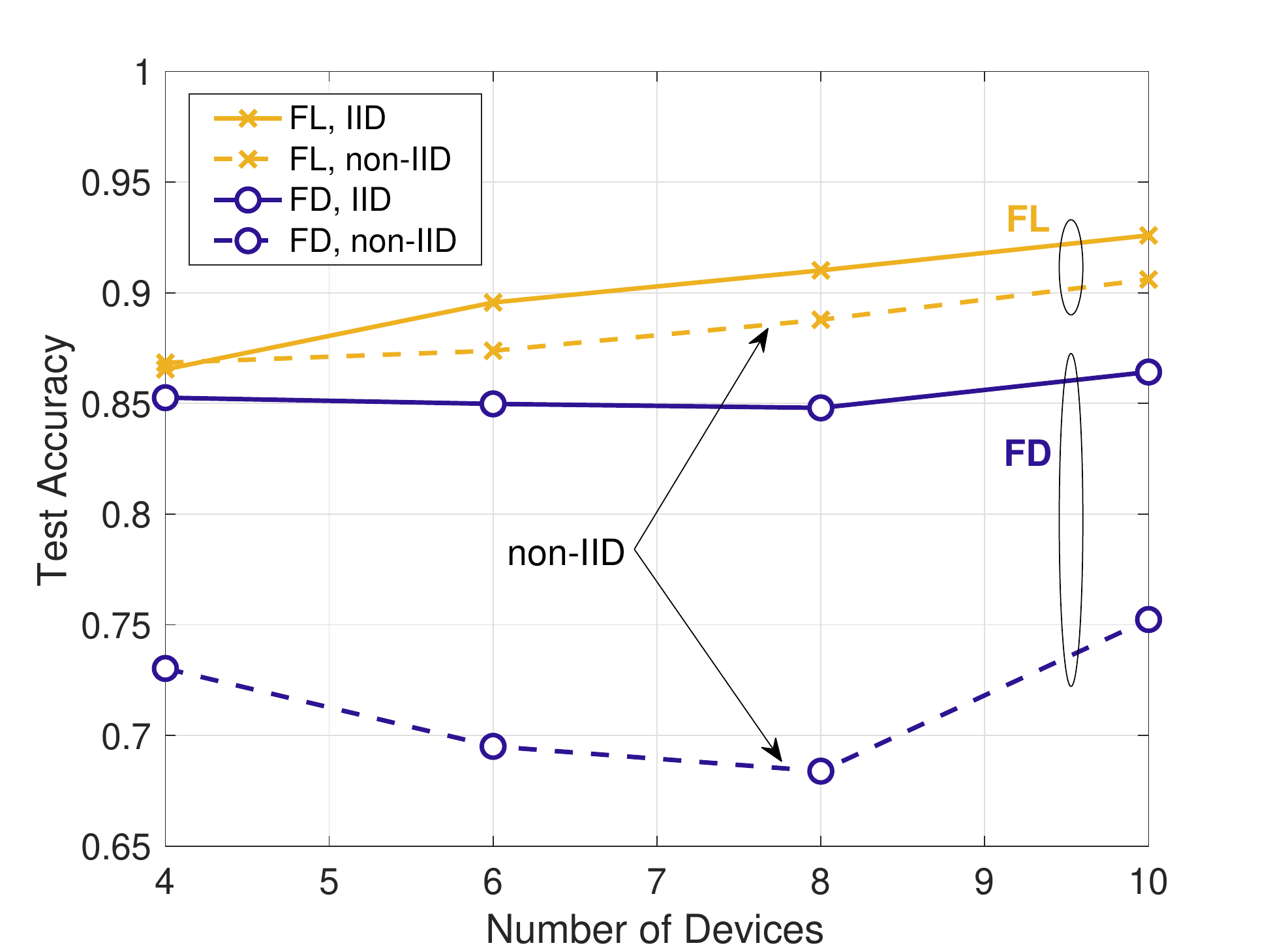}}
	\subfigure[Sum communication cost.]{\includegraphics[width=.85\linewidth]{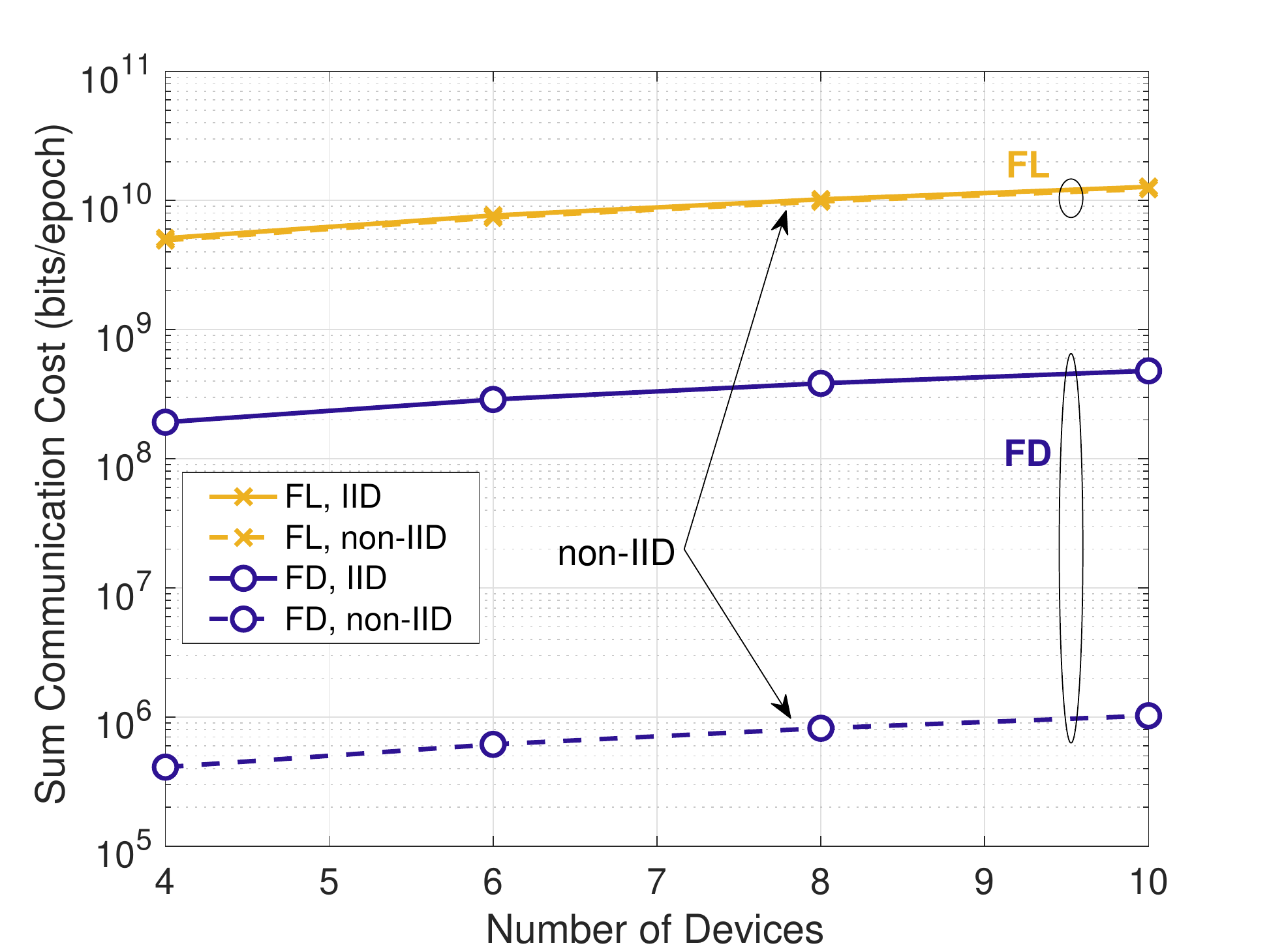}}
	\caption{Comparison between FD and FL in terms of (a) test accuracy and (b) sum communication cost of all devices per epoch, under an IID or non-IID MNIST data.} \label{FD_sim}
\end{figure}

\subsection{Representative Result}
To see the effectiveness of FD, we consider
the  MNIST (hand-written $0$-$9$ images)  classification task performed by $10$ devices.  Figure~\ref{FD_sim} illustrates the performance of FD for both cases of an IID local dataset and a non-IID dataset whose local data samples are imbalanced across labels. The result shows that for different numbers of devices, FD can always reduce around $10,\!000$x communication payload sizes per communication round compared to FL. Considering both fast convergence and payload size reduction, FD reduces the total communication cost until convergence by over $40,\!000$x compared to FL. Nonetheless, FD still comes at the cost of compromising accuracy, particularly under non-IID data distributions. 

\subsection{Summary and Research Opportunities} FD is a very efficient way of training models in a communication-efficient manner, which comes on par with the performance of federated learning. FD is still in its infancy and several interesting future directions include the co-design of wireless communication and federated distillation as opposed to treating them separately. Another natural extension is to investigate the cost-benefits of model quantization, distillation and their inherent trade-offs.

\section{Distributed Inference over Wireless Networks}\label{se:DI} 

While we have so far focused on the training aspect of ML at the wireless edge, another important component of edge learning is the inference stage. Once a ML model is trained using the available data, this model is then used to make inferences (classification or regression) on new data samples. In standard settings, training is considered to be the most computationally demanding phase of ML problems; and hence, most research has focused on improving the efficiency of distributed training; however, in the case of wireless edge devices and networks, inference is also challenging due to the power and complexity limitations of edge devices, and the latency requirements of the applications. This is particularly the case for inference using complex DNN models, whose dimensions can easily run into hundreds of millions. For example, the popular residual network architecture ResNet-50 image classification applications consists of 50 convolutional layers, and requires close to 100 megabytes of memory for storage and approximately 4 giga floating point operations (FLOPS) for each image. As an extreme example, the GPT-3 model trained for natural language processing has 175 billion parameters. Implementing such models on edge devices, especially within the time frames required by most edge applications is not feasible due to memory and computational limitations of edge devices. For example, in autonomous vehicles immediate detection of obstacles is critical to avoid accidents, putting stringent latency constraints on the inference task; however, the on-board processing units on a small drone may not be capable of carrying out large DNN inference in a timely manner. Similarly, data rates and processing speeds required in certain applications, such as particle physics experiments \cite{Duarte_2018} or in wireless communications  \cite{ramjee2019fast} can make online inference extremely challenging. In other scenarios, even when the processing capabilities of the devices or the data volume and latency constraints do not pose a significant challenge, it may not be possible to carry out inference locally if the data to make the inference on is distributed. For example, intelligence for surveillance applications may require images from multiple cameras, or in retrieval tasks, inference may require access to a database available in a remote server \cite{Jankowski:JSAC:21}. Similarly to federated training, communication becomes indispensable in such scenarios, and we need to guarantee that distributed inference can still be accomplished within the accuracy and latency constraints of the underlying application. With a few exceptions \cite{Jankowski:JSAC:21, Jankowski:SPAWC:20, Shao:ICC:20}, so far the physical layer aspects of distributed edge inference have been mostly ignored. 


\subsection{Fundamental Limits} 

From an information theoretic perspective, distributed inference over wireless channels can be treated as a joint source-channel coding version of remote rate-distortion problem \cite{Dobrushin:TIT:62, Wolf:IT:70}. Here, the label to be inferred can be considered as the required reconstruction based on the available data sample. The distortion measure can either be \textit{log loss} in the case of classification, or squared error distortion in the case of regression. Such an interpretation is followed in \cite{Sreekumar:IT:20} by further simplifying the inference task as a distributed binary hypothesis testing (HT) problem. Consider, for example, an \textit{observer}, that has independent identically distributed (i.i.d.) samples of a random variable. The observer can communicate over a noisy channel to a remote \textit{decision maker}, which wants to make a decision regarding the underlying probability distribution governing the data samples observed by the observer. Assuming a binary hypothesis testing scenario, e.g., the samples come from one of the two candidate distributions, it is shown in \cite{Sreekumar:ISIT:19} that \textit{separation} is optimal; that is, the simple scheme in which the observer locally performs optimal Neyman-Pearson test and communicates its decision to the tester using the best channel code for the two messages, achieves the best asymptotic error-exponent. 

Since we do not impose any computational constraints at the observer, this result is aligned with the intuition that the inference should be made locally at the observer as it has access to all the relevant data, and the implication of this result is that remote inference does not have a substantial impact on the performance, as long as the local decision can be conveyed to the remote decision maker reliably. However, the problem becomes significantly more challenging if both the observer and the decision maker have their own local observations, correlated with each other, and the goal is to make a decision on their joint distribution. In this scenario, since the observer has access only to its own observations, it cannot make a local decision no matter how much processing power it has; instead it must convey some features of its observations to help the decision maker to make the correct decision. 

It is known that when the goal of the decision maker is to reconstruct the observation of the observer within some distortion constraint, rather than deciding on their joint distribution, separate source-channel coding is asymptotically optimal \cite{Merhav:IT:03}; that is, it is optimal for the observer to first compress its observations into as few bits as possible satisfying the distortion constraint, and then to transmit these bits to the decision maker reliably using a capacity-achieving channel code. An interesting question here is whether such a scheme is still optimal when the goal is to make a decision on the joint distribution rather then reconstruction the source samples, such as the case in most ML problems. We note here that, even though hypothesis testing can also be viewed as a rate-distortion problem with a particular distortion metric, it is not an \textit{additive metric} as in standard rate-distortion problems, that is, the distortion between the original source vector and its reconstruction measured by the sum of the distortions between individual elements. It is shown in \cite{Sreekumar:IT:20} that the optimality of separation breaks down in the remote hypothesis testing problem. Interestingly, it is also shown in \cite{Sreekumar:IT:20} that the optimal error exponent can be achieved by a separation-based scheme for the special case of \textit{testing against independence}; that is, when deciding whether the samples at the observer and the decision maker come from a known distribution or are independent of each other. This result shows that communication and inference cannot be separated even in the asymptotic limit without loss of optimality.  On the other hand, how to design such joint schemes is mostly an open area of research.

\subsection{Neural Network Compression and Acceleration} 

From a practical point of view, since state-of-the-art performance is achieved by DNNs in most practical inference problems, the research has focused on implementing neural network inference on edge devices under the aforementioned constraints on the computational capabilities and memory of the devices, and the available power and bandwidth for communications. A possible approach to solve this problem is model architecture optimization, where the goal is to adjust the size and complexity of DNN architectures to the constraints of the edge device without sacrificing their performance. There are several approaches to achieve this in the literature. A more straightforward approach, similarly to those used for reducing the communication load during training, is to employ parameter pruning and quantization in order to reduce and remove redundant parameters that do not have a significant impact on the performance. It was discovered early on that pruning can reduce the network complexity and help address the overfitting problem \cite{Hanson:NIPS:89, LeCun:NIPS:90, Hassibi:NIPS:93}. Today there are many advanced pruning algorithms, and we refer the reader to \cite{liu2020pruning} for a detailed survey. Another effective approach is to impose sparsity constraints during training, through which we directly obtain a sparse network architecture, rather than trying to reduce a complex network to a sparse one through pruning  \cite{Lebedev_2016_CVPR, Wen:NeurIPS:16}. Network quantization, instead of removing some of the weights, tries to reduce the number of bits required to represent the network weights. In \cite{Gupta:JMLR:15, courbariaux2014training}, fixed-point representations are employed, and it is shown that very low precision is sufficient not just for inference based on trained networks but also for training them. Some works focused on training neural networks with only a single-bit binary weights \cite{Courbariaux:NIPS:15, Courbariaux:arXiv:16}, showing that DNNs can still perform well, significantly reducing their complexity and memory requirements. In \cite{lin2016neural}, by stochastically binarizing the weights the authors convert multiplications to sign changes, which further simplifies network operations. 

DNN compression can also be treated as a standard source compression problem, and vector quantization techniques can be employed for codebook-based compression to reduce the memory requirements. Hashing is used in \cite{Chen:ICML:15}, while \cite{Gong:arXiv:14} employed vector quantization. Huffman coding is applied in \cite{han2016deep} to further reduce the redundancy in quantized network weights.

\subsection{Joint Edge-Device Inference} 

While offloading data to the edge server for inference is one end of the spectrum, fully local inference using the above compression techniques can be considered as the other end. But, there can be a wide variety of solutions that lie in between, where we benefit from the local computation capabilities of the edge devices, but rather than carrying out full-fledged local inference, we also benefit from the edge servers, and the devices and the edge servers carry out inference tasks in a cooperative manner. A standard approach for edge-device cooperative inference is to split the DNN architecture into two, where the first several layers are carried out on the device, while the remainder are offloaded to the edge server. Such a distributed DNN arhitecture was first proposed in \cite{Teerapittayanon:ICDCS:17}, where a small DNN model is deployed on end devices while a larger NN model is employed in the cloud. For each inference query, the device first rapidly performs local inference using the local model for initial feature extraction, and even completes the inference if the model is confident based on the local features. Otherwise, the end device forwards the result of the local operations to the cloud, which performs further processing and final classification. This approach, by adaptively deciding on the offloading, provides a better use of the local resources and reduces the communication load compared to always offloading to the cloud, but also increases the accuracy compared to fully local inference. Moreover, since only the features that are required for the inference task are offloaded to the edge server, there is an inherent privacy protection as well.

In a parallel work, the neurosurgeon approach in \cite{Neurosurgeon} proposed joint computation between the edge device and the edge server by partitioning the layers between the two. By characterizing the per-layer execution time and the amount of data that needs to be conveyed to the edge server at the output of each layer, \textit{neurosurgeon} decides how to divide a complex DNN architecture between the device and the server. This approach is further extended in \cite{JointDNN} where the computations in a DNN are modeled as a directed acyclic graph, and the optimal computation scheduling between the edge device and server are studied for a large class of DNN architectures. It is shown in \cite{JointDNN} that for generative and autoencoder models, multiple data transfers between the device and the cloud may be required. 

It is observed in \cite{Neurosurgeon, JointDNN} that, in some DNN architectures, particularly those used for classification tasks, the data size at the output of the initial layers may be even larger than the input size. This would mean that carrying out the initial layers locally at the edge device might increase the communication cost. Lossless compression of the features using Portable Network Graphics (PNG) algorithm \cite{png} is considered in \cite{JointDNN}. Further reduction in the communication load can be achieved by using lossy compression. In \cite{edge_host_partitioning}, authors propose applying JPEG compression on the features before transmitting them to the edge server. On the other hand, standard image compression codecs have been designed for visual quality of the reconstructed image; and hence, they may remove high-frequency components of the features that are important for the classification task. In \cite{dnn_decoupling}, rather than employing standard compression codes, quantized feature maps are compressed using Huffmann coding. Alternatively, in the BottleNet architecture proposed in \cite{bottlenet}, a learnable feature reduction unit is introduced prior to JPEG compression to make sure only the most relevant features are compressed and forwarded to the edge server. 

More recently, in \cite{Shi:pruning, shao2020communicationcomputation, Jankowski:SPAWC:20} pruning techniques have been combined with DNN splitting to further reduce the computational load on the edge device. Thanks to pruning, more layers can be computed at the edge device within the latency and computational constraints. Pruning also provides a certain level compression by removing some of the less significant features. 

\subsection{Joint Edge-Device Inference Over a Wireless Channel} 

Above approaches abstract out the wireless channel as an error-free ideal bit-pipe, and focus only on the feature compression problem, ignoring the impacts of communication in terms of latency, complexity, or reliability. However, lossy transmission of feature vectors to the edge server over a wireless channel is essentially a \textit{joint source-channel coding (JSCC) problem}, and separation is known to be suboptimal under strict latency constraints imposed by inference problems \cite{ElGamalKim:book, Gunduz:IT:13, deepJSCC}. 

While JSCC has been studied extensively in the literature, past work mainly focus on the transmission of image or video sources, following a model-driven approach exploiting particular properties of the underlying source and channel statistics \cite{vanDyck:PIEEE:99, Cheung:ICIP:96, Tung:CL:18}. Recently, an alternative fully data-driven DNN-based scheme, called DeepJSCC, has been introduced \cite{deepJSCC, Kurka:JSAIT:20, kurka2020bandwidthagile}. DeepJSCC not only beats state-of-the-art image transmission schemes (e.g., BPG image compression + LDPC channel coding) in many scenarios, particularly in terms of perception sensitive quality measures (e.g., structured similarity index measure, SSIM), but also provides `graceful degradation' with channel quality, making it attractive for many edge inference applications where the ultra low latency requirements may render channel estimation infeasible. Note also that, these works only consider the latency and the complexity of the operations pertaining to the DNN layers, while ignoring those associated with channel coding/ decoding and modulation, which can be substantial, particularly if we want to operate at a rate close to the capacity of the underlying channel with low probability of error. DeepJSCC significantly reduces the coding/decoding delay compared to conventional digital schemes. Yet another advantage of employing DeepJSCC for edge inference is that, as opposed to conventional digital compression schemes like JPEG of BPG, DeepJSCC has the flexibility to adapt to specific source or channel domains through training. This makes DeepJSCC especially attractive for edge inference as we do not have compression schemes designed for generic feature vectors, whose statistics would change from application to application. 

A joint source-channel inference problem is first considered in \cite{jankowski2019deep, Jankowski:JSAC:21}, where a wireless image retrieval is studied. In the scenario considered, the image of a person captured by a remote camera is to be identified within a database available at an edge server. Here, the camera cannot make a local decision as it does not have access to the database. In \cite{Jankowski:JSAC:21}, two approaches are proposed, both employing DNNs for remote inference: a task-oriented DNN-based compression scheme for digital transmission and a DNN-based analog JSCC approach, similarly to the DeepJSCC scheme of \cite{deepJSCC}. It is observed that the proposed JSCC approach, which maps the feature vectors directly to channel inputs (no explicit compression or channel coding is carried out), performs significantly better. 

In \cite{Shao:ICC:20}, the authors employ JSCC for the joint mobile device- edge server inference problem, and called the new architecture BottleNet++, as this combines the approach in \cite{bottlenet} with DeepJSCC. A significant improvement in compression efficiency is achieved by BottleNet++ compared to directly transmitting the compressed feature vectors. In \cite{Jankowski:SPAWC:20}, pruning is employed jointly with DeepJSCC, and it is shown that an order of magnitude reduction in required channel bandwidth is possible compared to \cite{Shao:ICC:20}.

\section{Multi-Agent Reinforcement Learning over Wireless Networks}\label{se:ADL}

The previous sections introduced the supervised learning algorithms used for networking data analysis and inference, next, we introduce the RL for wireless network control and optimization.  

\subsection{Preliminaries of RL}

RL enables the wireless devices to learn the control and management strategies such as resource allocation schemes by interacting with their dynamic wireless environment \cite{4445757}. 
Next, we introduce three basic RL algorithms that are generally used for wireless networks.

\subsubsection{Single Agent RL} The formal model of a single agent RL can be described as a Markov decision process (MDP) \cite{4445757}.   Hence, the model of a single agent RL consists of four components: agent, state, action, and reward. The agent refers to the device that implements the RL algorithm. The state describes the environment observed by the agent at each time slot. A reward evaluates the immediate effect of an action given a state.

 Single agent RL enables the agent to find a policy that maximizes the expected discounted reward while only receiving the immediate reward at each learning step. During the single agent RL training process, the agent first observes its current state, and then performs an action. As a result, the agent receives its immediate
reward together with its new state. The immediate reward and new
state are used to update the agent's policy. This process
will be repeated until the agent finds a policy that can maximize the expected discounted reward.

In wireless networks, single agent RL algorithms can be considered as the centralized algorithms used for network control and optimization. In particular, single RL algorithms that are implemented by a central controller are mainly used for solving non-convex or time dependent optimization problems. For example, one can use single RL algorithms to optimize the trajectory of an unmanned aerial vehicle \cite{hu2020meta, 9014325, 8960453}. However, as the number of mobile devices that are considered by single agent RL increases, the action and state space of the single agent RL will significantly increase thus increasing the training complexity and decreasing the convergence speed. Meanwhile, as the number of the considered devices increases, the overhead of collecting state information of all devices increases which further increases the training complexity of single agent RL. Therefore, it is necessary to design distributed RL that can be jointly implemented by multiple devices.

\subsubsection{Independent Multi-Agent RL} Independent multi-agent RL is the simplest MARL algorithm. In the independent MARL, each device implements the single-agent RL individually. In consequence, each device aims to maximize its own expected discounted reward without considering other devices. Given the simple implementation, the agents that perform independent MARL does not need to share any RL information with other devices. Therefore, in wireless networks, independent MARL are generally used for the devices or the base stations (BSs) that cannot communicate with each other. Since the agents do not share any RL information, independent MARL are not guaranteed to converge and it also cannot find a local optimal solution to maximize the sum expected discounted reward of all agents.

\subsubsection{Collaborative Multi-Agent RL}  Collaborative MARL requires the agents to share some RL information with other agents. In particular, each agent can share its reward, RL model parameters, action, and state with other agents. For different collaborative RL algorithms, they may share different RL information. For example, the collaborative MARL designed in \cite{8395443} requires the agents to share their state and action information. In contrast, value decomposition network \cite{hu2020distributed} requires the agents to share their rewards. The training complexity and performance of a collaborative MARL algorithm depends on the information that each agent needs to share. The authors in \cite{StarCraft} had compared the training complexity and performance of different collaborative MARL algorithms. 

\subsection{State-of-the-Art}
Now, we discuss a number of recent works on the use of RL algorithms for network control and optimization. In particular, the authors in \cite{8395443} designed the recurrent neural network based RL algorithms for spectrum resource allocation. Using game theory, the designed RL algorithm is proved to converge to a mixed-strategy Nash equilibrium. In \cite{8714026}, the authors provided a comprehensive survey for the use of RL for solving wireless communication problems. The authors in \cite{chaccour2021seven} provided an introduction of using meta-learning based RL for Terahertz based wireless systems. The authors in \cite{6542770} designed an independent MARL algorithm for optimizing the spectral efficiencies of BSs and verified that collaborative MARL can achieve better performance than independent MARL. In \cite{9184075}, the authors designed a voting-based MARL algorithm that uses a primal-dual algorithm to find the optimal policy for large scale IoT systems. The authors in \cite{9277627} proposed a single-agent RL algorithm for optimizing the movement and transmit power of the
UAV, phase shifts of the reconfigurable intelligent surfaces (RIS), and the 
dynamic decoding order.

\subsection{Representative Result}

\begin{figure}[!t]
  \begin{center}
   \vspace{0cm}
    \includegraphics[width=9cm]{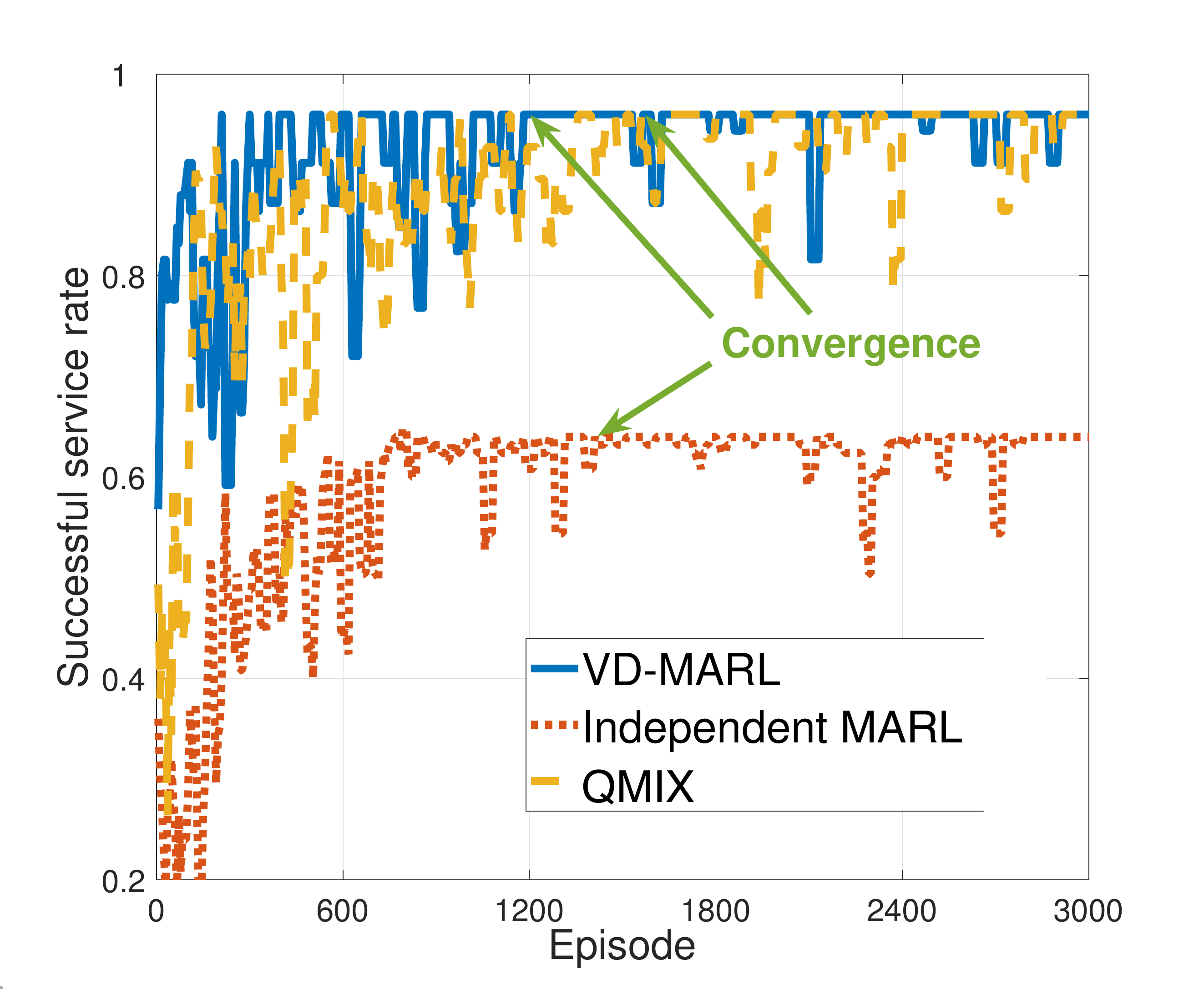}
    \vspace{-0.25cm}
    \caption{\label{VD-MARL} Convergence of the VD-MARL algorithm.}
  \end{center}\vspace{-0.8cm}
\end{figure}

One representative result on the development of MARL for UAV trajectory design can be found in the work \cite{hu2020distributed}. In the considered model, a team of UAVs is dispatched to cooperatively serve several clusters of ground users that have dynamic and unpredictable uplink access demands. The UAVs must cooperatively navigate to maximize coverage of the dynamic requests of the ground users. This trajectory design
problem is formulated as an optimization framework whose goal is to find optimal trajectories that maximize
the fraction of users served by all UAVs (called successful service rate hereinafter). Traditional optimization algorithms such as branch and bound are not suitable to solve this problem as the successful service rate achieved by each UAV is unpredictable due to the dynamic and unpredictable uplink access demands of ground users. 
Hence, we designed a novel MARL called VD-MARL that merges the concept of value decomposition network, model agnostic meta-learning, with the policy gradient
framework to optimize the trajectories of all UAVs. 
The proposed MARL algorithm enables each UAV to use the successful service rate achieved by all UAVs to estimate the expected successful service rate achieved by all UAVs over all states thus finding the local optimal trajectories for all UAVs. In particular, implementing VD-MARL, each UAV only needs to share its reward with other UAVs and hence, the overhead of RL information exchange among multiple UAVs significantly reduces.  

Fig. \ref{VD-MARL} shows the convergence of the VD-MARL algorithm. In this figure, we consider three RL algorithms: a) the proposed VD-MARL algorithm, b) the independent MARL algorithm based on actor critic, and c) QMIX in \cite{pmlrv80rashid18a}. From Fig. \ref{VD-MARL}, we can see that VD-MARL improve the successful service rate by up to 54\% compared to the independent MARL algorithm. This is because the
VD-MARL can find a team optimal strategy to maximize the successful service rate of all UAVs. The independent MARL algorithm, however, find a strategy that maximize each UAV's individual successful service rate. This figure also shows that VD-MARL improves the convergence speed by up to 31\% compared to the QMIX algorithm. This stems from the fact that the neural network in QMIX used to estimate the estimated future team reward remarkably increases the complexity of QMIX.

\subsection{Research Opportunities}

Using MARL for improving wireless network performance requires addressing a number of key problems including:

\subsubsection{Convergence Analysis} 
To analyze the optimality of RL solutions as well as the time and energy used for training RL algorithms, one important problem is to analzye the RL convergence. The existing works have used MDP to analyze the convergence of the single agent RL algorithms and game theory for simple MARL convergence analysis. However, none of these existing works can analzye the convergence of advance MARL algorithms such as QMIX due to complex RL information exchange and neural network updates. Therefore, in this problem, there are several issues including: 1) whether the studied MARL algorithm can find the optimal solution, 2) the number of iterations that MARL needs to converge, 3) how the number of MARL agents affects the convergence, 4) how approximation errors caused by ML models affects the MARL convergence.  

\subsubsection{Optimization of Wireless Networks for MARL Implementation} In wireless networks, the MARL convergence depends not only on the RL parameters such as the size of ML model but also on the wireless networking factors such as limited number of resource blocks (RBs), imperfect RL parameter transmission, and limited transmit power and computational power of devices. In particular, the number of RBs determines the number of devices that can perform MARL algorithm. Meanwhile, the dynamic wireless channels may cause errors on the transmitted RL parameters. In addition, the limited transmit power and computational power will significantly affect the time used for the RL model update and RL parameter transmission. Therefore, key problems in the implementation of MARL over wireless networks exists in many areas such as 1) optimization of RB allocation and device scheduling for RL parameter transmission, 2) reliable and energy efficient RL parameter transmission, 3) joint optimization of RL training methods and wireless resource allocation for minimizing RL convergence time, 4) coding and decoding method design, and 5) the deployment of advanced wireless techniques such as terahertz and intelligent reflecting surface.

\section{Conclusion}\label{se:conclusion}

In this paper, we have provided a comprehensive study of
the deployment of distributed learning over wireless networks. We have introduced four distributed learning frameworks, namely, federated learning, distributed inference, federated distillation, and multi-agent reinforcement learning. For each learning framework, we have introduced the motivation for deploying it over wireless networks. Meanwhile, we have presented a detailed literature review, an illustrative example, and future research opportunities for each distributed learning framework. Such an in-depth
study on the deployment of distributed learning over wireless networks provides the guidelines for optimizing, designing, and operating distributed learning based wireless communication systems.

\bibliographystyle{IEEEbib}
\bibliography{references,AirComp_Ref}
\end{document}